\documentclass[journal]{IEEEtran}
\usepackage{amsmath,amsfonts}
\usepackage{algorithmic}
\usepackage{algorithm}
\usepackage{array}
\usepackage[caption=false,font=normalsize,labelfont=sf,textfont=sf]{subfig}
\usepackage{textcomp}
\usepackage{stfloats}
\usepackage{url}
\usepackage{verbatim}
\usepackage{graphicx}
\usepackage{cite}
\usepackage{bm}
\usepackage[colorlinks,bookmarksopen,bookmarksnumbered,citecolor=blue, linkcolor=blue, urlcolor=blue]{hyperref}
\usepackage{booktabs}
\usepackage{multirow}
\usepackage{colortbl}
\usepackage{makecell}
\usepackage{bbding}
\usepackage{soul}
\usepackage{nicematrix}
\begin{document}

\title{Binarized High-Efficiency RAW Video \\ Restoration and Beyond}

\author{Tianyu Zhu, Ying Fu, \IEEEmembership{Senior Member,~IEEE}, Hesong Li, Gengchen Zhang, Xin Yuan, \IEEEmembership{Senior Member,~IEEE}, and Yulun Zhang, \IEEEmembership{Member,~IEEE}
        % <-this % stops a space
\thanks{This work was supported by Beijing Natural Science Foundation (L253009), the National Natural ScienceFoundation of China (62331006), and the Fundamental Research Funds for the Central Universities. (\textit{Corresponding author: Ying Fu.})}
\thanks{Tianyu Zhu, Ying Fu, and Hesong Li are with School of Computer Science and
Technology, Beijing Institute of Technology, Beijing, China. Email: zhutianyu@bit.edu.cn, fuying@bit.edu.cn, lihesong2@bit.edu.cn.}% <-this % stops a space
\thanks{Gengchen Zhang is with Chinese Aeronautical Establishment, Beijing, China. Email:
zxmyx2@163.com.}
\thanks{Xin Yuan is with Westlake University, Hangzhou, China. Email:
xyuan@westlake.edu.cn.}
\thanks{Yulun Zhang is with MoE Key Lab of Artificial Intelligence, AI
Institute, Shanghai Jiao Tong University, Shanghai, China. Email:
yulun100@gmail.com.}
\thanks{The code is available at https://github.com/Tony1882880/BinRVR.}
}

% The paper headers
\markboth{IEEE Transactions on Pattern Analysis and Machine Intelligence,~Vol.~XX, No.~X, Month~2026}%
{Shell \MakeLowercase{\textit{et al.}}: A Sample Article Using IEEEtran.cls for IEEE Journals}

\maketitle

\begin{abstract}
RAW video restoration is fundamental to high-quality low-level perception and serves as the basis for a wide range of downstream vision applications. While binary neural networks (BNNs) enable efficient lightweight deployment for image enhancement, their deficiencies in modeling temporal coherence and activation value distributions hinder their effectiveness when applied to video scenarios. In this paper, we propose BinRVR, a binarized RAW video restoration framework that reduces computation and parameters by approximately 96\% while incurring only about 4\% performance degradation. Specifically, we present a Binarized Information Interaction Module (BIIM) to jointly model spatial and temporal information in an efficient and unified manner. Moreover, we develop a Distribution-Aware Binarized Convolution (DAB-Conv) that leverages the statistics of full-precision activations to mitigate quantization errors. The proposed framework further supports multi-bit quantization, enabling flexible accuracy-efficiency trade-offs across different hardware constraints. Extensive experiments demonstrate that our BinRVR achieves competitive performance compared with state-of-the-art binarized methods on RAW video restoration tasks, including low-light enhancement, denoising, deblurring, and super-resolution. We further explore the potential of our method on downstream video applications, including object detection and monocular depth estimation.
\end{abstract}


\section{Introduction}
\IEEEPARstart{V}{ideos} captured in real-world scenarios inevitably suffer from quality degradation due to sensor limitations and challenging environmental conditions, resulting in low illumination, noise, motion blur, and low resolution. Such degradations not only impair human visual perception but also undermine the performance of downstream vision tasks, including object detection \cite{det1,det_survey}, depth estimation \cite{depth1,lyp2}, and segmentation \cite{seg_survey,seg1}. To address these challenges, a large body of research has been devoted to video restoration, aiming to recover high-quality videos from degraded inputs.

Existing video restoration methods can be broadly divided into two categories according to the domain of the input and output videos, \textit{i.e.}, RGB-based methods \cite{VRT}, \cite{ShiftNet}, and RAW-based methods \cite{Real-RawVSR}, \cite{LLRVD}. Since RAW data bypass the nonlinear processing of the image signal processor (ISP), they more faithfully preserve scene irradiance and retain original image information \cite{LLRVD}, \cite{raw1}, \cite{raw2}. Unlike RGB frames that undergo irreversible ISP operations such as white balance, tone mapping, and denoising, RAW data maintain a nearly linear sensor response and a wider dynamic range, which are crucial for recovering fine structures and low-light details. Therefore, RAW videos are more suitable as both input and output representations to maximally preserve information.

\begin{figure}[t]
    \centering
    \includegraphics[width=\columnwidth]{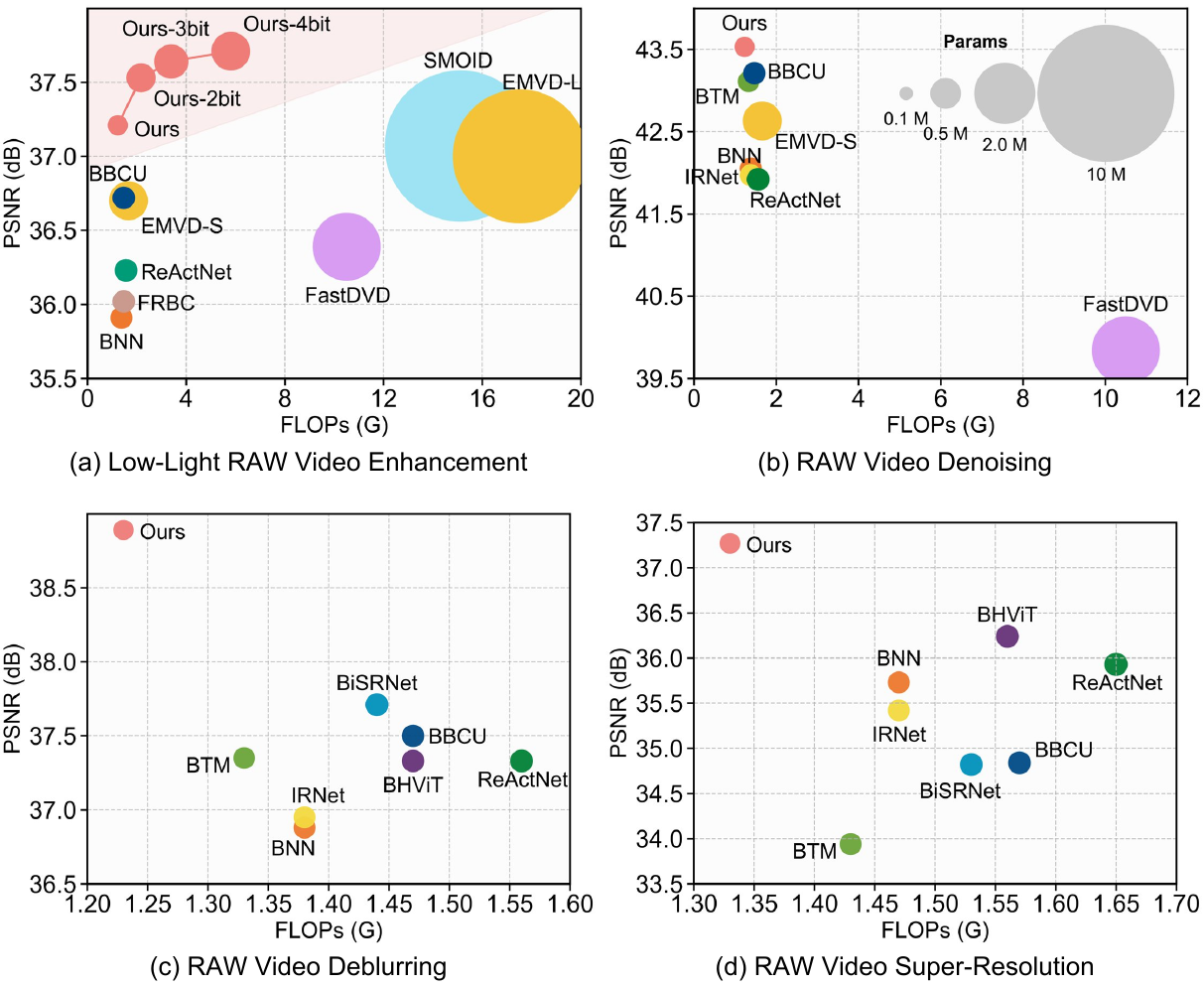}
    \vspace{-7mm}
    \caption{Performance and model complexity comparison with state-of-the-art binary neural networks (BNNs) and lightweight full-precision methods in RAW video restoration. The size of each circle reflects the model’s parameter count. Our method outperforms both the BNN-based methods and lightweight full-precision methods with comparable model complexity.}
    \vspace{-4mm}
    \label{teaser}
\end{figure}

Deep neural networks (DNNs) have achieved remarkable performance in RAW video restoration \cite{VRT,ShiftNet,FloRNN}. However, their high computational cost and large parameter size hinder deployment on resource-constrained devices such as smartphones, unmanned aerial vehicles (UAVs), and Internet of Things (IoT) platforms. To address this limitation, various acceleration and compression techniques have been explored, including pruning \cite{NetworkSlimming,prunning1}, lightweight full-precision architecture design \cite{MobileNetV2}, knowledge distillation \cite{FitNets}, quantization \cite{LSQ}, and binarization \cite{BNN}. These techniques improve efficiency from different perspectives and are not mutually exclusive. Among them, network quantization has attracted increasing attention due to its broad applicability across architectures and its ability to reduce computation and storage. Binary Neural Networks (BNNs) represent an extreme form of quantization, where both weights and activations are constrained to 1-bit values, typically represented as $+1$ and $-1$. When supported by appropriate hardware kernels, binarization enables compact storage and bitwise computation, making it a promising direction for efficient RAW video restoration.

Despite their efficiency, applying BNNs to RAW video restoration remains challenging. First, most existing studies have concentrated on tasks such as classification and single-image restoration \cite{BBCU}, whereas inter-frame relationships in BNN-based video restoration have received little attention. Second, many state-of-the-art video restoration models \cite{VRT,deformable2} rely on complex operations such as multi-head attention, optical flow estimation, and deformable convolutions to capture spatio-temporal features. These operations are not naturally friendly to BNNs, where weights and activations are represented by signs, \textit{i.e.}, $+1$ and $-1$. Sign-based features discard magnitude and fine ordering information. For attention modules, quantized Vision Transformer studies show that attention maps and softmax-related nonlinear operations often require special quantization-aware designs \cite{FQViT,IViT}. For optical-flow-based alignment, reliable flow estimation depends on discriminative feature matching and continuous displacement prediction \cite{RAFT,optical_flow2}, while deformable convolution predicts input-dependent offsets to determine continuous sampling locations \cite{deformable,deformable2}. Inferring these similarities, flows, or offsets from coarse binary cues can shift sampling locations and amplify alignment errors during feature aggregation. Third, BNNs suffer from internal discrepancies, as the distributions of weights and activations vary across layers, channels, frames, and degradation types \cite{XNOR-Net}, making uniform sign-based binarization prone to substantial quantization errors.

In this paper, we propose a \textbf{Bin}arized High-Efficiency \textbf{R}AW \textbf{V}ideo \textbf{R}estoration model (\textbf{BinRVR}) to address the above challenges. Different from existing binarized restoration methods that mainly focus on image-level tasks, BinRVR provides a systematic binarized solution for multi-task RAW video restoration, including low-light enhancement, denoising, deblurring, and super-resolution. Specifically, BinRVR employs a unidirectional recurrent structure that enables efficient inference with minimal memory overhead, making it particularly suitable for deployment on resource-limited devices. To address the severe representational constraints imposed by binarized RAW video restoration, we present a Binarized Information Interaction Module (BIIM) for joint modeling of temporal dependencies and spatial structures. BIIM integrates a temporal shift-and-extend operation with grouped strip-shaped spatial convolutions, enabling efficient spatiotemporal interaction with negligible overhead and full compatibility with binarized and low-bit quantized models. To further reduce quantization error, we develop a Distribution-Aware Binarized Convolution (DAB-Conv), which employs channel attention to derive real-valued scaling factors from activations, thereby incorporating full-precision distribution information. Extensive experiments demonstrate that our BinRVR substantially reduces computation and parameters with only minor performance degradation, and achieves a favorable accuracy-efficiency trade-off compared with state-of-the-art binarized methods and representative lightweight full-precision models under comparable complexity, as shown in Fig.~\ref{teaser}.

To summarize, our main contributions are as follows:
\begin{itemize}
\item We propose BinRVR, a high-efficiency binarized framework for multi-task RAW video restoration, explicitly extending binarization from image-level restoration to temporally coherent RAW video restoration;
\item We introduce a Binarized Information Interaction Module (BIIM) that enables efficient joint modeling of inter-frame temporal dependencies and intra-frame spatial structures through lightweight and binarization-friendly operations;
\item We develop a Distribution-Aware Binarized Convolution (DAB-Conv) that leverages activation distribution information to mitigate quantization errors and naturally supports multi-bit quantization for flexible accuracy-efficiency trade-offs under hardware constraints.
\end{itemize}

The remainder of this paper is organized as follows. Section \ref{related} reviews related works on video restoration and binary neural networks. Section \ref{method} describes the proposed BinRVR framework in detail, including the BIIM and DAB-Conv. Section \ref{experiment} reports extensive experimental results, including quantitative evaluations, visual comparisons, and discussions. Finally, Section \ref{conclusion} concludes the paper.

\section{Related Works} \label{related}
In this section, we review related work in two parts. We first discuss video restoration methods, especially those involving RAW data, and then review BNNs with quantization.

\subsection{Video Restoration}
Video restoration encompasses tasks such as low-light video enhancement, video denoising, video deblurring, and video super-resolution. Existing methods can be broadly categorized into single-task methods \cite{recurrent1,BasicVSR++,FloRNN,SFIN,zhangyingkai1,zhangyingkai2,lhs-small,zhangyingkai3} and multi-task frameworks \cite{VRT,ShiftNet,TriMSOD}, \cite{recurrent3}. In the following, we review representative methods from two perspectives, namely overall architecture design and spatio-temporal feature fusion.

\vspace{1mm}\noindent\textbf{Network architecture design.}
Early DNN-based video restoration methods typically took all frames of a video as input to the model \cite{SMOID}, \cite{RViDeNet}, but this design is unsuitable for variable-length videos and incurs high storage costs. Recent methods mainly adopt either multi-frame sliding-window architectures \cite{sli_win1}, \cite{deformable2}, \cite{LLRVD}, \cite{FastDVD}, \cite{MMNet} or single-frame recurrent structures \cite{ShiftNet}, \cite{recurrent1}, \cite{BasicVSR++}, \cite{FloRNN}, \cite{EMVD}. Chen \textit{et al.} \cite{MMNet} fused features from a sliding window with single-frame spatial features and produced multi-frame output. Chan \textit{et al.} \cite{BasicVSR++} enhanced the bidirectional recurrent framework with additional skip connections. To combine the advantages of both, Liang \textit{et al.} \cite{VRT} proposed a single-frame-to-single-frame architecture with a global flow estimation module. Despite these advances, sliding-window methods are effective at capturing short-term dependencies but are less effective in modeling long-term relations, whereas recurrent structures leverage hidden states for long-term modeling but may be limited in short-term correlations and are prone to error accumulation.

\begin{figure*}[t]
    \centering
    \includegraphics[width=\textwidth]{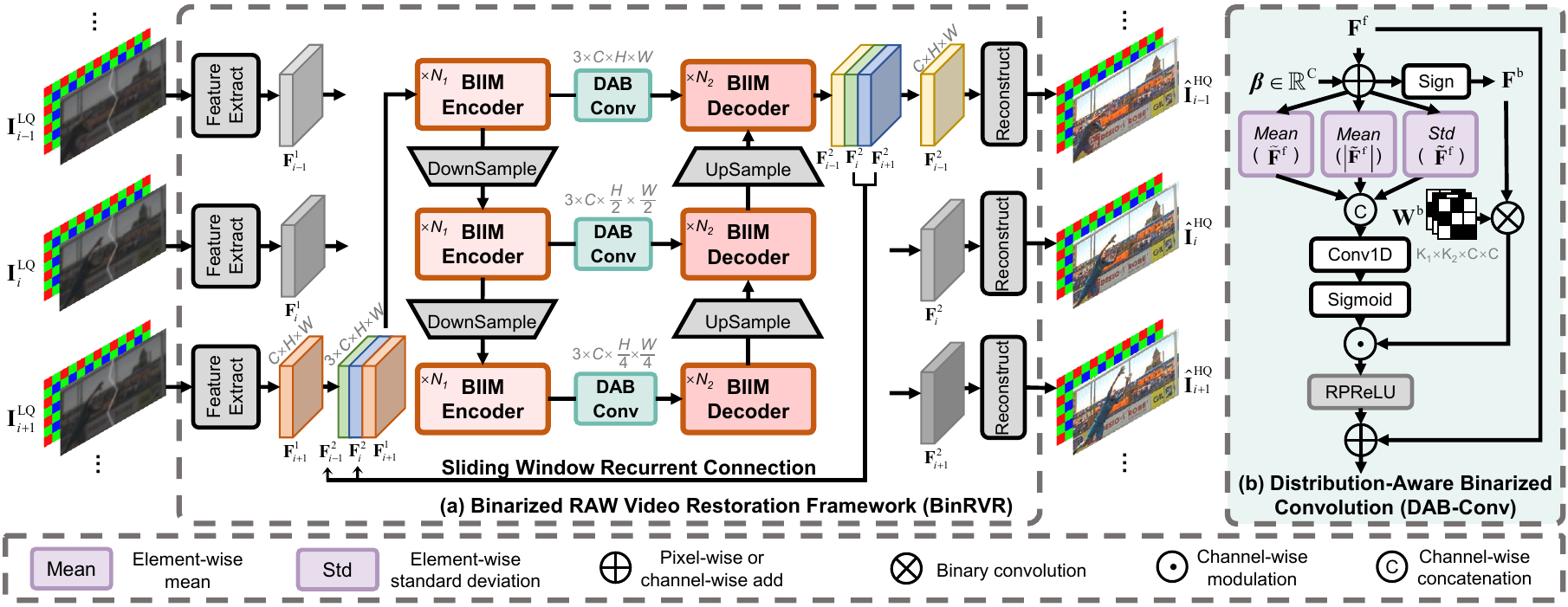}
    \vspace{-6.5mm}
    \caption{\textbf{Illustration of the proposed BinRVR.} (a) Overall architecture of our binarized RAW video restoration framework. The recurrent arrows indicate the propagated features between adjacent sliding windows. Efficient inter-frame and intra-frame information interaction under binarization is achieved by the proposed Binarized Information Interaction Module (BIIM), whose detailed structure is shown in Fig. \ref{framework2}. (b) Distribution-Aware Binarized Convolution (DAB-Conv), which leverages feature distribution information to reduce quantization error and improve representation capability in binarized networks.}
    \vspace{-3mm}
    \label{framework}
\end{figure*}

\vspace{1mm}\noindent\textbf{Spatio-temporal feature interaction.}
Unlike image restoration \cite{SwinIR,FCDFusion,tcsvt1,CJE1,CJE2,CJE3,CJE4,liuqiankun,wangbinfeng}, which mainly focuses on spatial information, video restoration additionally requires effective extraction and fusion of temporal cues. Some methods \cite{optical_flow1}, \cite{optical_flow2}, \cite{VRT} explicitly align or warp frames using optical flow estimation. Others adopt deformable convolutions \cite{deformable}, \cite{CRVD}, \cite{deformable2} or attention mechanisms \cite{VRT}, \cite{IART}, \cite{attention1} to implicitly fuse spatio-temporal features across frames. However, these approaches incur high computational and memory costs, and their performance degrades noticeably under quantization, especially binarization. To reduce complexity, lightweight fusion strategies have been explored. Li \textit{et al.} \cite{ShiftNet} aggregated multi-frame features effectively by applying group-wise spatial and temporal shifts within feature channels. Yue \textit{et al.} \cite{Demoiréing} proposed a RAW video demoiréing method that leverages channel-separated features and spatial modulation to exploit the complementary properties of RAW RGGB channels, while aligning multi-scale temporal features. Despite these efforts, lightweight spatio-temporal feature fusion remains underexplored in the context of quantization and binarization.

\subsection{Binary Neural Networks and Quantization}
Binary neural networks (BNNs) are proposed as an extreme case of quantization to exploit the efficiency of bitwise operations on hardware \cite{BNN}, where both weights and activations are represented with only 1 bit. However, such extreme quantization inevitably leads to performance degradation. To mitigate this issue, extensive efforts have focused on reducing quantization errors \cite{XNOR-Net,XNOR-Net++,quan_func1}, refining loss functions \cite{ReActNet,LCR}, approximating gradients \cite{Bireal,IRNet}, and optimizing training strategies \cite{train_stg1,train_stg2}. Recent methods further improve binary models through optimization-friendly binarization functions and high-accuracy binary architectures, such as BiPer \cite{BiPer} and BNext \cite{BNext}. Most studies validate the effectiveness of BNNs on high-level vision tasks such as image classification and object detection \cite{bnn_det1,scal_fac1}.

By contrast, applications of BNNs to low-level vision remain limited and mainly focus on image restoration \cite{BBCU,BTM}. Representative works include BBCU \cite{BBCU}, which emphasizes refining residual connections, batch normalization, and activation functions, Advanced BNN \cite{AdvancedBNN} and Rectified Binary Network \cite{Rectified-BSR} for binarized single-image super-resolution, and BHViT \cite{BHViT}, which explores a hybrid convolution and Transformer design. These methods demonstrate the potential of binarization for image restoration, but they mainly focus on RGB-domain single-image tasks. Binarized RAW video restoration additionally requires efficient temporal modeling and robustness to RAW-domain activation distribution variations, which are the main focus of this work.

Compared with BNNs, multi-bit quantization provides less aggressive compression but generally yields better performance. Several studies have investigated multi-bit quantization for image super-resolution. QuantSR \cite{QuantSR} introduced a low-bit framework with a learnable redistributive quantizer to improve information representation. It further employs a dynamic quantization architecture to adaptively adjust network depth. Hong \textit{et al.} \cite{AdaBM} proposed AdaBM, an adaptive framework that adjusts bit-widths according to image complexity and layer sensitivity. Nevertheless, extensions of quantization and binarization to video restoration remain scarce.

\section{Binarized RAW Video Restoration} \label{method}
In this section, we formalize the RAW video restoration task and elaborate on the motivation for constructing a binarized model that balances computational efficiency and restoration quality. We then present the overall architecture of the proposed BinRVR framework, followed by a detailed description of the Binarized Information Interaction Module (BIIM) and the Distribution-Aware Binarized Convolution (DAB-Conv). Finally, we introduce a multi-bit compatible design for high-efficiency RAW video restoration.

\begin{figure*}[t]
    \centering
    \includegraphics[width=1.0\textwidth]{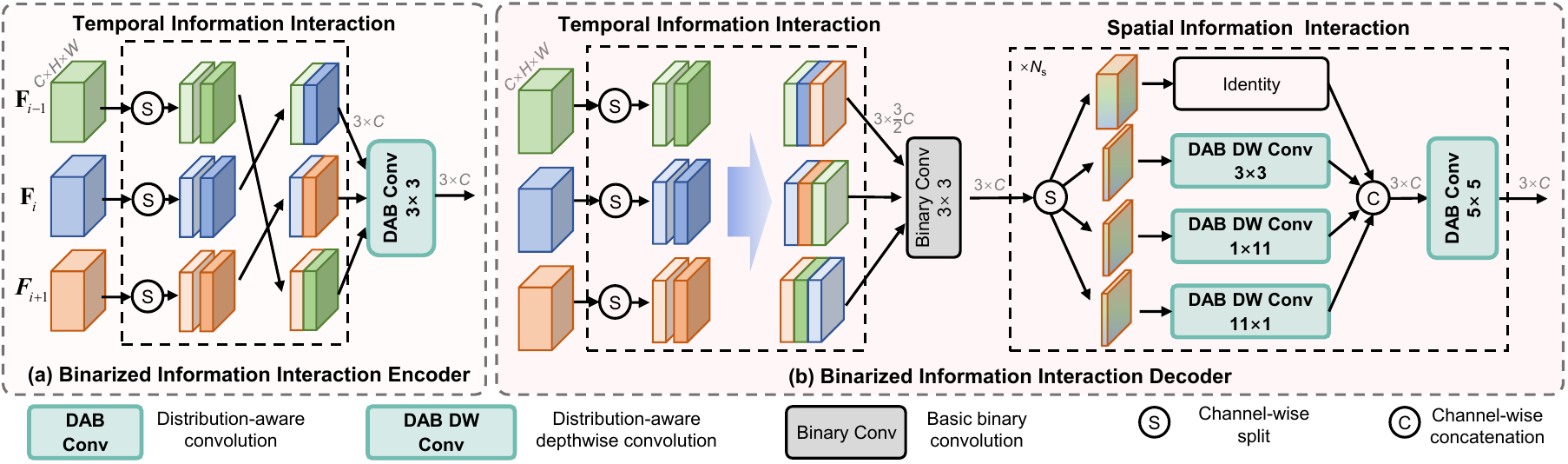}
    \vspace{-7mm}
    \caption{\textbf{Detailed illustration of the proposed Binarized Information Interaction Module (BIIM).} It is designed to enable efficient interaction between inter-frame temporal information and intra-frame spatial information under binarization through lightweight operations. (a) Binarized Information Interaction Encoder. (b) Binarized Information Interaction Decoder.}
    \vspace{-2mm}
    \label{framework2}
\end{figure*}

\subsection{Formulation and Motivation}
We outline the formulations of RAW video restoration, quantization, and binarization. We then introduce the motivations of our BIIM and DAB-Conv.

\vspace{1mm}\noindent\textbf{Formulation of RAW video restoration.} The task of RAW video restoration can be formulated as the transformation of an input sequence of low-quality (LQ) RAW video frames \( \{\mathbf{I}_1^{\text{LQ}}, \mathbf{I}_2^{\text{LQ}}, \dots, \mathbf{I}_T^{\text{LQ}}\} \), where each frame \( \mathbf{I}_i^{\text{LQ}} \) is a Bayer pattern RAW image with spatial resolution \( H \times W \) and packed into a tensor of size \( \mathbf{I}_i^{\text{LQ}} \in \mathbb{R}^{4 \times H/2 \times W/2} \). The pixel values of the RAW images are first uniformly normalized to the range \([0, 1]\) based on the camera-specific black level and white level. The output is a sequence of high-quality (HQ) RAW video frames \( \{\hat{\mathbf{I}}_1^{\text{HQ}}, \hat{\mathbf{I}}_2^{\text{HQ}}, \dots, \hat{\mathbf{I}}_T^{\text{HQ}}\} \). As shown in Fig.~\ref{framework}\textcolor{blue}{(a)}, the process is expressed as
\begin{equation}
\{\hat{\mathbf{I}}_1^{\text{HQ}}, \hat{\mathbf{I}}_2^{\text{HQ}}, \dots, \hat{\mathbf{I}}_T^{\text{HQ}}\} = \mathcal{G}\left(\{\mathbf{I}_1^{\text{LQ}}, \mathbf{I}_2^{\text{LQ}}, \dots, \mathbf{I}_T^{\text{LQ}}\}; \mathbf{\Theta}\right),
\end{equation}
where \( \mathcal{G} \) represents the restoration model and \( \boldsymbol{\Theta} \) denotes the learnable parameters of the model. The restoration process aims to recover high-quality details and maintain temporal consistency across the video sequence.

\vspace{1mm}\noindent\textbf{Formulation of binarization and quantization.}
We denote $b$ as the quantization bit-width. Binary neural networks (BNNs) represent the extreme case of quantization under the single-bit setting ($b=1$), where weights and activations are constrained to $\{+1,-1\}$. The binarization mapping is written as
\begin{equation}
\mathcal{B}(x)
= \text{Sign}(x)
= \begin{cases}
+1, & x\ge 0,\\
-1, & x<0,
\end{cases}
\end{equation}
with $\mathcal{B}(\cdot)$ denoting the binarization function. Storing $+1$ as \texttt{1} and $-1$ as \texttt{0} enables hardware-level acceleration. Multiplication is replaced by XNOR, and accumulation by bitcount \cite{xnor, XNOR-Net}. The binarized convolution is computed as
\begin{equation}
\mathbf{F}^\text{b} \otimes\mathbf{W}^\text{b}
= \text{Bitcount}\big(\text{XNOR}(\mathbf{F}^\text{b},\mathbf{W}^\text{b})\big).
\end{equation}

For $b\ge2$, multi-bit quantization maps full-precision values to a broader discrete set than binarization. With clipping range $[x_{\min}, x_{\max}]$, we use the following uniform quantizer, \textit{i.e.},
\begin{equation}
\mathcal{Q}(x) =
\left\lfloor
\frac{\text{Clamp}(x, x_{\min}, x_{\max}) - x_{\min}}{\delta}
\right\rceil\cdot\delta + x_{\min},
\end{equation}
where $\left\lfloor\cdot\right\rceil$ denotes rounding to the nearest integer and $\delta$ is the quantization step size.

\vspace{1mm}\noindent\textbf{Motivation of binarized information interaction module.}
Modeling interaction between inter-frame temporal information and intra-frame spatial structures is crucial for video restoration, yet remains challenging under binarization. Existing methods often rely on cross-attention for temporal modeling \cite{VRT}, resulting in high computational overhead. Others attempt to enlarge the receptive field through downsampling \cite{EVDR}, sacrificing spatial details. Techniques such as optical flow \cite{optical_flow1} and deformable convolution \cite{deformable} can further enhance spatio-temporal modeling, but they rely on discriminative real-valued feature matching or continuous offset prediction, which becomes fragile when features are constrained to $+1/-1$ signs. Inspired by the efficiency-oriented designs of \cite{ShiftNet}, \cite{InceptionNeXt}, these limitations motivate a binarized module for efficient spatio-temporal interaction in RAW videos.

\vspace{1mm}\noindent\textbf{Motivation of distribution-aware binarized convolution.}
Binarized convolutions are sensitive to activation distributions, as sign-based quantization discards magnitude information and introduces quantization errors \cite{ReActNet}. Existing binarization methods \cite{BNN,IRNet,BBCU} employ either per-layer or per-channel scaling strategies, and many representative methods estimate scaling factors mainly from the mean absolute value of weights or activations, which captures the average magnitude but provides limited distribution information. For RAW video restoration, activation distributions vary across channels, frames, and degradation conditions. Our DAB-Conv extracts multiple distribution descriptors, including the mean, absolute mean, and standard deviation, and adaptively fuses them to predict input-dependent channel-wise activation scaling factors. These complementary statistics help adapt activation scaling to complex RAW-domain feature distributions and reduce quantization-induced degradation during restoration.

\subsection{Overall architecture design}

The overall architecture of the proposed BinRVR is illustrated in Fig.~\ref{framework}\textcolor{blue}{(a)}, which shows the restoration process for three consecutive RAW frames \(\mathbf{I}_i^{\text{LQ}}\) from a video sequence. The framework consists of three stages. In the feature extraction stage, a binarized U-Net \cite{UNet} extracts frame-wise features such as \(\mathbf{F}_{i}^{1}\). In the recursive encoding-decoding stage, a multi-scale encoder-decoder network processes the stacked features with downsampling by factors of two and four to capture both global structures and fine details. Each scale integrates \(N_1\) binarized information interaction encoders and \(N_2\) binarized information interaction decoders, as shown in Fig.~\ref{framework2}, and detailed in Section~\ref{BIIM}. This stage produces refined features \(\mathbf{F}_{i}^2\). In the reconstruction stage, \(\mathbf{F}_{i}^2\) is used to generate the high-quality frame \(\hat{\mathbf{I}}_{i}^{\text{HQ}}\), while partial features of this iteration are forwarded to the next iteration via the sliding window recurrent connection, which will be described in detail. As illustrated in Fig.~\ref{framework}\textcolor{blue}{(b)}, the distribution-aware binarized convolution is a core module throughout all stages, binarizing both feature maps and convolutional kernels to significantly reduce computational cost and memory usage. Further details are provided in Section~\ref{DAB-Conv}.

\vspace{1mm}\noindent\textbf{Sliding window recurrent connection. }
Conventional video restoration methods predominantly employ either a multi-frame sliding window framework \cite{sli_win1}, \cite{deformable2}, or a single-frame recurrent architecture \cite{ShiftNet}, \cite{recurrent1}. While the former effectively captures short-term dependencies, it fails to exploit long-term temporal relationships; the latter leverages hidden state propagation for long-term modeling but is limited in handling short-term correlations and is prone to cumulative errors. 

\begin{figure}[t]
    \centering
    \includegraphics[width=\columnwidth]{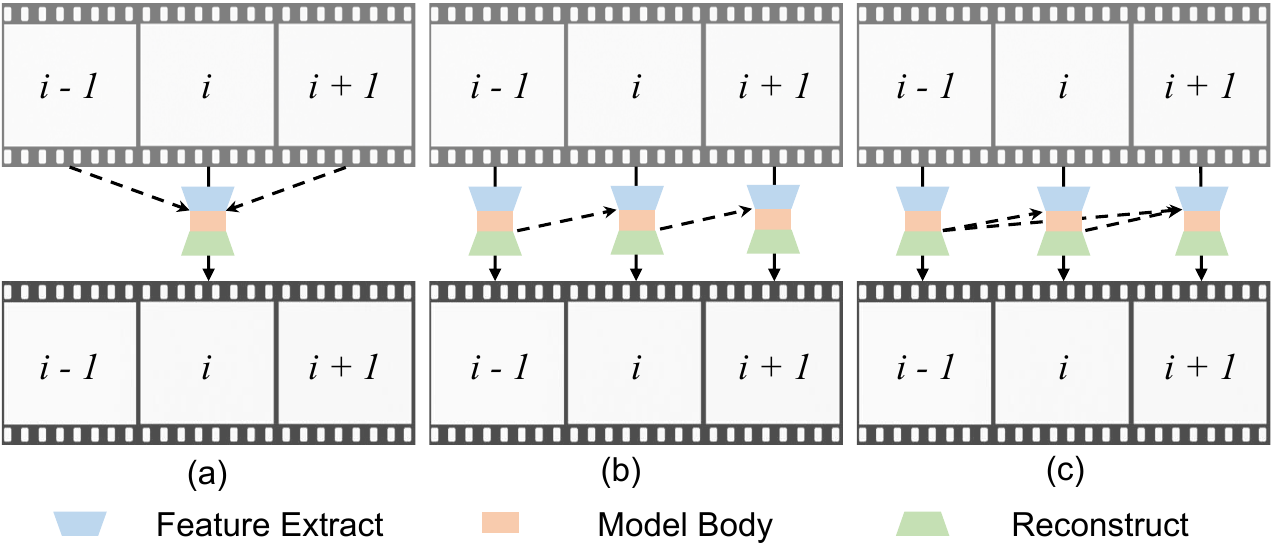}
  \vspace{-7mm}
    \caption{Comparison of video restoration architectures. (a) Multi-frame sliding window framework. (b) Single-frame recurrent architecture. (c) Proposed sliding window recurrent connection.}
    \vspace{-2mm}
    \label{architecture}
\end{figure}

To overcome these limitations, we propose the sliding window recurrent connection, which integrates the strengths of both paradigms. As depicted in Fig.~\ref{architecture}, this design preserves the sliding window structure while incorporating recurrent connections across adjacent windows, thereby enabling efficient short-term feature aggregation and long-term dependency modeling. The input at time step \(t\) of the recursive encoding-decoding stage can be expressed as
\begin{equation}
    \mathbf{F}^1(t) = \mathcal{C}\left(\mathbf{F}_i^2(t-1), \mathbf{F}_{i+1}^2(t-1), \mathbf{F}_{i+1}^1(t)\right),
\end{equation}
where $\mathcal{C}(\cdot)$ denotes channel-wise concatenation, and $\mathbf{F}$ denotes corresponding features.
Our sliding window recurrent connection offers a principled and computationally efficient framework for enhancing temporal information flow.

\vspace{2mm}\noindent\textbf{Loss function. }
We use the joint loss function of Charbonnier loss \cite{charbonnier} and structural similarity index measure loss between the predicted HQ frames \( \hat{\mathbf{I}}^{\text{HQ}} \) and the ground-truth HQ frames \( \mathbf{I}^{\text{HQ}} \) as
\begin{equation}
    \mathcal{L} = \mathcal{L}_\text{c}(\hat{\mathbf{I}}^{\text{HQ}}, \mathbf{I}^{\text{HQ}}) + \lambda \mathcal{L}_\text{s}(\hat{\mathbf{I}}^{\text{HQ}}, \mathbf{I}^{\text{HQ}}),
\end{equation} 
where \(\mathcal{L}_\text{c}\) represents the Charbonnier loss \cite{charbonnier}, \(\mathcal{L}_\text{s}\) represents the SSIM loss, and \(\lambda\) is a task-dependent hyperparameter. We set \(\lambda=0.01\) for low-light enhancement, denoising, and super-resolution, and \(\lambda=0.1\) for deblurring, where structural similarity is particularly important for recovering blurred RAW details in deblurring.

\subsection{Binarized information interaction module} \label{BIIM}
 The Binarized Information Interaction Module (BIIM) is designed to enlarge temporal and spatial receptive fields in RAW video restoration. As illustrated in Fig.~\ref{framework2}, BIIM consists of a temporal and a spatial enhancement module. The temporal module introduces no additional learnable parameters or convolutional operations and only incurs negligible arithmetic overhead through channel splitting and concatenation, while the spatial module employs lightweight strip-shaped depthwise convolutions compared with conventional large-kernel binarized depthwise convolutions. We note that the normalized FLOPs and parameter analysis mainly reflects arithmetic and storage complexity, while wall-clock latency may also depend on memory access and hardware implementation. To balance efficiency and effectiveness, only the temporal module is used in the encoder, whereas both temporal and spatial modules are applied in the decoder.

\vspace{1mm}\noindent\textbf{Temporal information interaction.} To facilitate improved information fusion among adjacent frames (current, previous, and subsequent), we introduce the temporal information interaction module in the BIIM decoder, as illustrated in Fig.~\ref{framework2}\textcolor{blue}{(b)}. A simplified version of this module is employed in the BIIM encoder, as illustrated in Fig.~\ref{framework2}\textcolor{blue}{(a)}. The module takes as input three feature maps $\{\mathbf F_{i-1}, \mathbf F_{i}, \mathbf F_{i+1}\} \in \mathbb{R}^{C \times H \times W}$, each of which is initially split into two parts along the channel dimension, formulated as
\begin{equation}
    \bigl[\mathbf f_{i+k}^1,\ \mathbf f_{i+k}^2\bigr]
    = \mathcal{S}(\mathbf F_{i+k}), \quad k \in \{-1,0,1\},
\end{equation}
where $\mathbf f \in \mathbb{R}^{C/2 \times H \times W}$ are the two partitioned segments of the feature map for each frame.

To ensure that each frame's feature map, after temporal enhancement, incorporates information from these three frames, the output feature map can be represented as 
\begin{equation}
    \mathbf F^{\rm T}_{\,i+k}
    = \mathcal{C}\bigl(\mathbf f_{i+k}^1,\ \mathbf f_{i+k+1}^2,\ \mathbf f_{i+k-1}^\alpha\bigr),
    \quad k\in\{-1,0,1\},
\end{equation}
where $\mathbf F^{\rm T} \in \mathbb{R}^{\frac{3}{2}C \times H \times W}$ denotes the output of the temporal-enhanced module, and $\alpha \in \{1,2\}$. This operation effectively expands the receptive field across frames without introducing additional learnable parameters or convolutional operations, making it suitable for binarization and quantization.

\begin{figure}[t]
    \centering
    \includegraphics[width=\columnwidth]{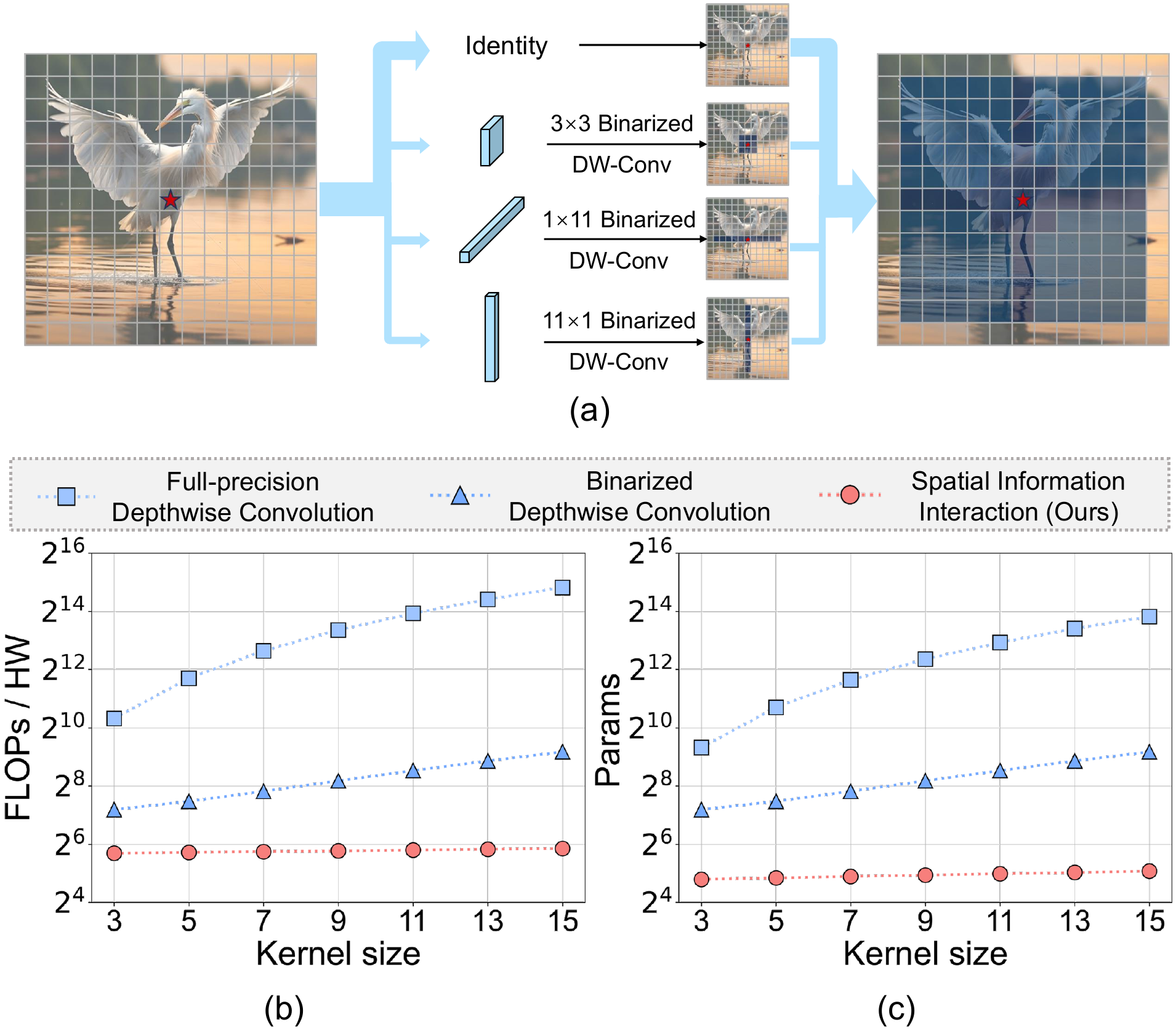}
     \vspace{-7mm}
    \caption{Illustration of spatial information interaction module.
(a) Illustration of the spatial information interaction module, where feature channels are split into identity, $3\times3$, $1\times11$, and $11\times1$ binarized depthwise convolution branches.
(b) FLOPs per spatial location versus kernel size.
(c) Parameter complexity versus kernel size.}
    \label{fig:sii}
    \vspace{-3mm}
\end{figure}

\vspace{1mm}\noindent\textbf{Spatial information interaction. }
The spatial information interaction module is designed to enlarge the receptive field in a lightweight and binarization-friendly manner for RAW video restoration. Large-kernel depthwise convolutions expand the receptive field and can improve performance \cite{ERF}, \cite{ConvNeXt}. However, excessively large square kernels significantly increase computation and parameters. Prior studies also suggest that processing only subsets of channels in depthwise layers can achieve competitive performance \cite{ShuffleNetv2}. Motivated by this, we replace large square kernels with horizontal and vertical strip-shaped kernels (\textit{e.g.}, \(1 \times 11\) and \(11 \times 1\)), each applied to a subset of channels. Meanwhile, another subset is directly propagated via identity mapping, while the remaining subset is processed with small square kernels (\textit{e.g.}, \(3 \times 3\)) through binarized depthwise convolutions as shown in Fig.~\ref{fig:sii}\textcolor{blue}{(a)}. Formally,
\begin{equation}
\begin{aligned}
&[\mathbf f_{\rm i}, \mathbf f_{\rm s}, \mathbf f_{\rm h}, \mathbf f_{\rm v}] = \mathcal{S}(\mathbf F^{\rm T}),\\
&\tilde{\mathbf F}^{\rm S} = \mathcal{C}(\mathbf f_{\rm i},
    \mathcal{D}_{3 \times 3}(\mathbf f_{\rm s}),
    \mathcal{D}_{1 \times 11}(\mathbf f_{\rm h}),
    \mathcal{D}_{11 \times 1}(\mathbf f_{\rm v})
),
\end{aligned}
\end{equation}
where \(\mathbf{F} ^{\rm T}\) is the output of our temporal information interaction module, and \(\mathbf f_{\rm i}, \mathbf f_{\rm s}, \mathbf f_{\rm h}, \mathbf f_{\rm v}\) denote subsets processed by identity mapping, small square kernels, horizontal strip kernels, and vertical strip kernels, respectively. \(\mathcal{D}(\cdot)\) denotes the distribution-aware binarized depthwise convolution, which will be introduced in Section \ref{DAB-Conv}. Finally, the spatial information
interaction module applies a \(5\times5\) distribution-aware binarized depthwise convolution on the intermediate feature map \(\tilde{\mathbf F}^{\rm S}\) to obtain the final output \(\mathbf F^{\rm S}\).

Following the commonly adopted efficiency estimation protocol for binarized networks, we weight binary operations as $1/64$ of their floating-point FLOPs and binary parameters as $1/32$ of their full-precision counterparts, while keeping bias terms in full precision. Under this setting, the normalized computational cost of our spatial information interaction module (per spatial location) is
\begin{equation}
\frac{\mathrm{FLOPs}}{HW}
= 2\left[\frac{1}{64}\left(\frac{C_{\rm in}\,K_{\rm p}}{8}\times 2
+ \frac{C_{\rm in}\,K_{\rm q}}{8}\right)
+ \frac{C_{\rm in}}{8}\times 3 \right],
\end{equation}
and the corresponding parameter complexity is
\begin{equation}
\mathrm{Params}
= \frac{1}{32}\left(\frac{C_{\rm in}\,K_{\rm p}}{8}\times 2
+ \frac{C_{\rm in}\,K_{\rm q}}{8}\right)
+ \frac{C_{\rm in}}{8}\times 3 ,
\end{equation}
where $K_{\rm p}$ denotes the kernel length of the strip-shaped convolutions, and $K_{\rm q}$ denotes the kernel size of the small square-shaped convolution. The last term in both equations corresponds to the full-precision bias of the three convolutional branches. Notably, the FLOPs and parameters of SII increase linearly with the kernel size, whereas those of full-precision and conventional binarized depthwise convolutions grow quadratically. Detailed comparisons are provided in Fig.~\ref{fig:sii}\textcolor{blue}{(b)} and Fig.~\ref{fig:sii}\textcolor{blue}{(c)}.

\subsection{Distribution-Aware Binarized Convolution} \label{DAB-Conv}
Binarized convolutions often suffer performance degradation due to the loss of full-precision details in weights and activations. A common remedy is to introduce real-valued scaling factors \cite{XNOR-Net, scaling_factor}. In full-precision networks, most weights approximately follow a symmetric bell-shaped distribution. We compute the weight scaling factors using the $L_1$-norm of each kernel, expressed as
\begin{equation}
\mathbf S^{\rm w}
= \frac{1}{C_{\rm in} K_1 K_2}
\Bigl[
\|\mathbf W^{\rm f}_1\|_1,\,
\|\mathbf W^{\rm f}_2\|_1,\,
\cdots,\,
\|\mathbf W^{\rm f}_{C_{\rm out}}\|_1
\Bigr],
\end{equation}
where $\mathbf W^{\rm f}_i \in \mathbb{R}^{C_{\rm in} \times K_1 \times K_2}$ is the full-precision kernel of the $i$-th output channel,
$\|\cdot\|_1$ denotes the $L_1$-norm computed as the sum of absolute values over all elements in the kernel,
and $\mathbf S^{\rm w} \in \mathbb{R}^{C_{\rm out}}$ denotes the channel-wise weight scaling factors used to mitigate binarization error.
For distribution-aware binarized depthwise convolutions, this formulation degenerates to the case $C_{\rm in}=1$.

Unlike weights, activations depend on input data and exhibit highly diverse and often asymmetric distributions. Simply reusing the weight scaling strategy is insufficient for robust binarization. To address this, we design distribution-aware activation scaling factors that explicitly incorporate statistical characteristics of the input. As illustrated in Fig.~\ref{framework}\textcolor{blue}{(b)}, a per-channel learnable bias $\boldsymbol{\beta}\in\mathbb{R}^C$ is first added to the real-valued activations to compensate for asymmetric distributions, yielding adjusted activations $\tilde{\mathbf F}^{\rm f} = \mathbf F^{\rm f} + \boldsymbol{\beta}$.

We then compute channel-wise descriptive statistics of $\tilde{\mathbf F}^{\rm f}$, including the mean, mean of absolute values, and standard deviation (all computed per channel), \textit{i.e.},
\begin{equation}
\begin{aligned}
&\tilde{\mathbf S}^{\rm a}
= \mathcal{C}\Bigl(\text{Mean}(\tilde{\mathbf F}^{\rm f}),\ \text{Mean}(|\tilde{\mathbf F}^{\rm f}|), \text{Std}(\tilde{\mathbf F}^{\rm f})\Bigr), \\[3pt]
&\mathbf S^{\rm a}
= \text{Sigmoid}\!\left(\text{Conv1D}(\tilde{\mathbf S}^{\rm a})\right),
\end{aligned}
\end{equation}
where $\tilde{\mathbf S}^{\rm a} \in \mathbb{R}^{3C}$ is the intermediate statistical descriptor and $\mathbf S^{\rm a} \in \mathbb{R}^C$ is the final activation scaling factor. Together, the above two steps define a distribution-aware scaling function $\Phi(\cdot)$ that maps the adjusted activations $\tilde{\mathbf F}^{\rm f}$ to channel-wise scaling factors.

Intuitively, DAB-Conv adapts the quantization scale to the activation distribution of each channel. The mean reflects the center shift of the distribution, the mean of absolute values reflects the average response magnitude, and the standard deviation reflects the dynamic range and dispersion. We do not claim that these descriptors are theoretically optimal in an information-theoretic sense; instead, they are lightweight and stable first- and second-order statistics that are closely related to binarization error. The lightweight Conv1D adaptively fuses these complementary cues to predict input-dependent channel-wise scaling factors, so that the quantized activations better match the full-precision ones.

After obtaining both weight and activation scaling factors, the distribution-aware binarized convolution is computed as
\begin{equation}
\begin{aligned}
&\mathbf F^{\rm b} = \text{Sign}(\tilde{\mathbf F}^{\rm f}),\\[3pt]
&\mathbf Y
= \text{RPReLU}\Bigl((\mathbf F^{\rm b} \otimes \mathbf W^{\rm b}) \odot \mathbf S^{\rm w} \odot \mathbf S^{\rm a}\Bigr)
+ \mathbf F^{\rm f},
\end{aligned}
\label{eq:bin_final}
\end{equation}
where $\mathbf Y$ is the final output, and $\text{RPReLU}(\cdot)$ denotes a learnable real-valued activation function following \cite{ReActNet}.

\subsection{Distribution-Aware Multi-Bit Quantization} \label{DAB-Q}

Although binarization offers substantial efficiency gains, real-world deployment often requires multiple quantization bit-widths to match hardware constraints and balance efficiency and accuracy. Multi-bit quantization better preserves representational fidelity with $b$-bit discrete values, but activation distributions still vary noticeably across layers and channels, and fixed ranges may cause biased scaling and mismatched dynamic ranges. 

To address this, we introduce our distribution-aware mechanism in the multi-bit setting, enabling adaptive scaling that matches the local activation statistics. A lightweight per-channel bias $\boldsymbol{\beta}$ is applied to align activation distributions, producing the adjusted activations $\tilde{\mathbf F}^{\rm f}$. The quantization range is then dynamically adapted through
\begin{equation}
    \mathbf{S}^{\rm a} = \Phi(\tilde{\mathbf F}^{\rm f}),
\end{equation}
where $\Phi(\cdot)$ summarizes activation statistics (channel-wise mean, absolute mean, and standard deviation over spatial locations) and maps them into channel-wise scaling factors.

Based on this adaptive scaling, multi-bit activation quantization is computed as $\mathbf F^{q} = \mathcal{Q}(\tilde{\mathbf F}^{\rm f}/\mathbf{S}^{\rm a})\cdot\mathbf{S}^{\rm a}$, where $\mathcal{Q}(\cdot)$ denotes the previously defined uniform quantizer and is not repeated here. Finally, the distribution-aware multi-bit convolution with residual connection is
\begin{equation}
    \mathbf{Y}
= \gamma\,\mathbf{F}^{\rm f}
+ \text{ReLU}\bigl((\mathbf{F}^{q} \otimes \mathbf{W}^{q})\odot \mathbf{S}^{\rm a}\bigr),
\end{equation}
where $\gamma$ is a learnable scalar, inspired by the residual balancing strategy in QuantSR \cite{QuantSR}, to balance the shortcut and quantized branches. This scalar is introduced only for multi-bit quantization, where the quantized branch retains richer discrete representations and its contribution can be adaptively balanced with the shortcut branch. For 1-bit binarization, Eq. \ref{eq:bin_final} already employs RPReLU, residual connection, and distribution-aware scaling to compensate for severe representation loss, making an additional balancing scalar less beneficial.

In contrast to standard multi-bit quantization, which applies fixed scaling across channels, our distribution-aware mechanism produces quantized features that retain robustness under diverse activation statistics and alleviate range mismatch across layers and inputs.

\vspace{1mm}\noindent\textbf{Gradient approximation.}
The sign and round functions used in binarization and quantization are non-differentiable, making direct back-propagation infeasible. Following previous works~\cite{Bireal,BBCU}, we adopt straight-through estimators (STEs) during training to approximate the gradients of these discrete operations. Specifically, for binarized weights, the gradient of the sign function is approximated by the derivative of a clip function, while for binarized activations, a smoother piecewise quadratic surrogate is used in the backward pass. For multi-bit quantization, the rounding operation is also optimized with STE, where the forward pass performs discretization and the backward pass approximates its gradient through the quantized branch. These approximations allow stable gradient propagation during training while keeping the inference path binarized or quantized.

\begin{table*}[t]
\centering
\footnotesize
\setlength{\tabcolsep}{0.7mm}
\renewcommand{\arraystretch}{1.12}
\caption{Quantitative comparison on SMOID \cite{SMOID} and LLRVD \cite{LLRVD} for low-light RAW video enhancement. For binarized methods, the best and second-best results are shown in \textbf{bold} and \underline{underlined}, respectively. Lightweight full-precision methods are additionally included for reference, with the best result shown in \textbf{bold}.}
\vspace{-2mm}
\definecolor{lightgray}{gray}{0.94} % 定义浅灰色
\definecolor{lightblue}{RGB}{220, 240, 255}
\begin{NiceTabular}{l|c|c||ccc|ccc|ccc||ccc}%四个c代表有四列且内容居中
\toprule

\multirow{3}{*}{Method} & \multirow{3}{*}{\makecell[c]{Params\\(M)$\downarrow$}} & \multirow{3}{*}{\makecell[c]{FLOPs\\(G)$\downarrow$}} & \multicolumn{9}{c||}{SMOID \cite{SMOID}} & \multicolumn{3}{c}{\multirow{2}{*}{LLRVD \cite{LLRVD}}} \\
\cline{4-12}
& & & \multicolumn{3}{c|}{Gain0} & \multicolumn{3}{c|}{Gain15} & \multicolumn{3}{c||}{Gain30} \\

& & & PSNR$\uparrow$ & SSIM$\uparrow$ & ST-RRED$\downarrow$ & PSNR$\uparrow$ & SSIM$\uparrow$ & ST-RRED$\downarrow$ & PSNR$\uparrow$ & SSIM$\uparrow$ & ST-RRED$\downarrow$ & PSNR$\uparrow$ & SSIM$\uparrow$ & ST-RRED$\downarrow$\\
\midrule 
Linear Scaled & - & - & 31.26 & 0.7665 & 0.4965 & 32.35 & 0.8108 & 0.2541 & 32.27 & 0.8358 & 0.4156 & 30.04 & 0.6941 & 0.0863\\
\midrule
\multicolumn{15}{l}{\textit{\textbf{Full-precision} methods}} \\
SMOID \cite{SMOID} & 12.21 & 15.11 & 39.74 & 0.9733 & 0.0446 
& 39.43 & 0.9751 & 0.0461 & 40.06 & 0.9769 & 0.0431 &
37.07 & 0.9580 & 0.0391 \\

RViDeNet \cite{RViDeNet} & 8.57 & 65.77 & 41.09 & 0.9778 & 0.0510 & 40.94 & 0.9803 & 0.0494 & 41.60 & 0.9818 & 0.0500 & 37.39 & 0.9620 & 0.0400\\

FastDVD \cite{FastDVD} & 2.48 & 10.50 & 40.37 & 0.9724 & 0.0765 & 40.66 & 0.9768 & 0.0664 & 40.59 & 0.9763 & 0.1812 & 36.39 & 0.9450 & 0.0724\\

EMVD-L \cite{EMVD} & 9.60 & 17.52 & 41.10 & 0.9785 & 0.0657 & 40.34 & 0.9804 & 0.0691 & 41.13 & 0.9819 & 0.0796 & 37.00 & 0.9576 & 0.0473\\

EMVD-S \cite{EMVD} & \textbf{0.81} & \textbf{1.66} & 39.71 & 0.9707 & 0.0732 & 39.23 & 0.9735 & 0.0804 & 39.88 & 0.9756 & 0.0875 & 36.70 & 0.9527 & 0.0527\\

LLRVD \cite{LLRVD} & 6.29 & 46.14 & 41.51 & 0.9799 & 0.0388 & 41.75 & 0.9823 & 0.0330 & 42.13 & 0.9840 & 0.0350 & 37.74 & 0.9650 & 0.0347\\

FloRNN \cite{FloRNN} & 10.49 & 24.57 & 41.39 & 0.9801 & 0.0468 & 40.74 & 0.9823 & 0.0560 & 41.55 & 0.9842 & 0.0558 & 37.47 & 0.9634 & 0.0377\\

ShiftNet \cite{ShiftNet} & 13.38 & 32.87 & \textbf{42.11} & \textbf{0.9836} & \textbf{0.0328} & \textbf{42.28} & \textbf{0.9848} & \textbf{0.0273} & \textbf{42.70} & \textbf{0.9863} & \textbf{0.0280} & \textbf{37.87} & \textbf{0.9661} & \textbf{0.0346} \\

\midrule
\multicolumn{15}{l}{\textit{\textbf{Binarized} methods}} \\
BNN \cite{BNN} & 0.26 & 1.38 & 38.34 & 0.9558 & 0.1078 & 38.30 & 0.9636 & 0.1191 & 38.75 & 0.9628 & 0.1045 & 35.91 & 0.9362 & 0.0669 \\

Bireal \cite{Bireal} & 0.26 & 1.38 & 38.38 & 0.9530 & 0.1138 & 38.50 & 0.9672 & 0.1086 & 38.74 & 0.9628 & 0.1013 & 36.22 & 0.9449 & 0.0591\\

IRNet \cite{IRNet} & 0.26 & 1.38 & 38.21 & 0.9517 & 0.1270 & 38
53 & 0.9647 & 0.1100 & 38.65 & 0.9634 & 0.1139 & 35.90 & 0.9348 & 0.0670\\

ReActNet \cite{ReActNet} & 0.28 & 1.56 & 38.80 & 0.9567 & 0.1049 & 38.71 & 0.9702 & 0.1140 & 39.23 & 0.9667 & 0.0961 & 36.23 & 0.9443 & 0.0608 \\

BTM \cite{BTM} & \underline{0.25} & \underline{1.33} & 39.58 & 0.9704 & 0.0824 & 39.43 & 0.9736 & 0.0879 & 39.77 & 0.9755 & 0.0986 & 36.64 & 0.9528 & 0.0532 \\

BBCU \cite{BBCU} & 0.27 & 1.47 & 39.63 & 0.9710 & 0.0793 & 39.97 & 0.9743 & 0.0635 & 40.16 & 0.9765 & 0.0730 & 36.72 & 0.9541 & 0.0523\\

BiSRNet \cite{BiSRNet} & 0.28 & 1.44 & 39.71 & 0.9708 & 0.0766 & 39.98 & 0.9741 & 0.0610 & 40.17 & 0.9764 & 0.0777 & 36.76 & 0.9548 & 0.0509\\

FRBC \cite{FRBC} & 0.26 & 1.38 & 31.35 & 0.7356 & 0.6570 & 32.88 & 0.8065 & 0.3950 & 32.97 & 0.7944 & 0.3994 & 36.02 & 0.9405 & 0.0626 \\

BHViT \cite{BHViT} & 0.27 & 1.47 & 39.87 & 0.9717 & 0.0752 & 40.05 & 0.9751 & 0.0621 & 40.53 & 0.9774 & 0.0703 & 36.70 & 0.9540 & 0.0527 \\

BRVE \cite{BRVE} & 0.30 & 1.49 & \underline{40.05} & \underline{0.9742} & \underline{0.0639} & \underline{40.25} & \underline{0.9765} & \underline{0.0557} & \underline{40.64} & \underline{0.9786} & \underline{0.0570} & \underline{37.07} & \underline{0.9581} & \underline{0.0455} \\

\rowcolor{lightgray}
\textbf{BinRVR (Ours)} & \textbf{0.22} & \textbf{1.23} & \textbf{40.25} & \textbf{0.9758} & \textbf{0.0625} & \textbf{40.45} & \textbf{0.9781} & \textbf{0.0523} & \textbf{40.83} & \textbf{0.9800} & \textbf{0.0554} & \textbf{37.21} & \textbf{0.9599} & \textbf{0.0436} \\

\bottomrule
\end{NiceTabular}
\vspace{-3mm}
\label{low-light-quantitative}
\end{table*}

\begin{figure*}[t]
    \centering
    \includegraphics[width=1.0\linewidth]{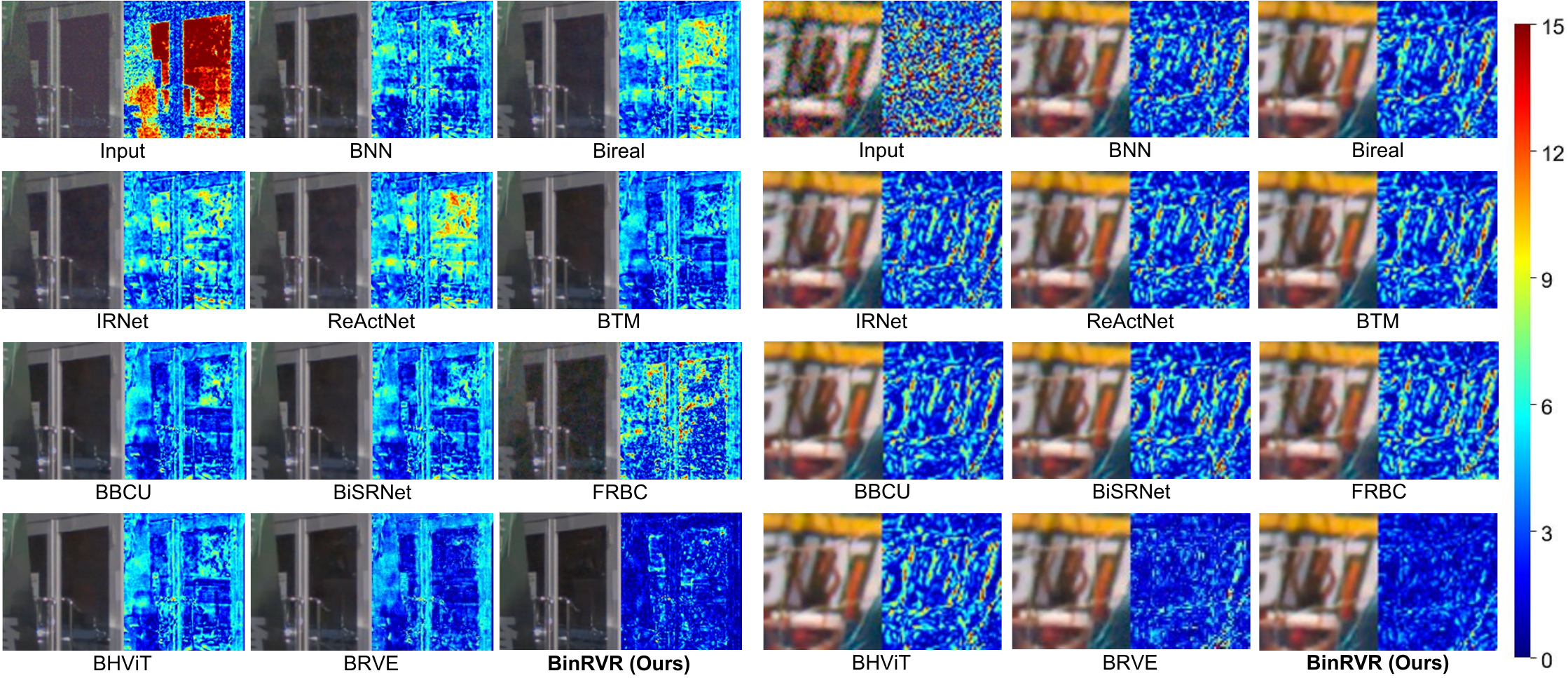}
     \vspace{-7mm}
    \caption{Visual comparisons of binarized methods on the SMOID~\cite{SMOID} and LLRVD~\cite{LLRVD} datasets for low-light RAW video enhancement (left: SMOID; right: LLRVD). For each example, the left half shows zoomed-in RGB results obtained by applying an ISP pipeline to the RAW outputs, while the right half presents the corresponding error maps with respect to the ground truth.}
    \label{fig:vis_ll}
    \vspace{-3mm}
\end{figure*}

\section{Experiments} \label{experiment}
In this section, we evaluate our BinRVR on four RAW video restoration tasks, \textit{i.e.}, low-light RAW video enhancement, RAW video denoising, RAW video deblurring, and RAW video super-resolution. We further evaluate the method under the multi-bit quantization setting on low-light enhancement, followed by discussions and downstream task evaluations.

\subsection{Experiment Settings}
We first describe the implementation details, followed by the binarized comparison methods and evaluation metrics, and finally present the model efficiency evaluation.

\vspace{1mm}\noindent\textbf{Implementation details.}
The models are trained on 10-frame sequences from all datasets, with each batch consisting of a single $256 \times 256$ RAW patch. The total batch size is set to 2. Data augmentation techniques such as random cropping, random horizontal flipping, random vertical flipping, and random rotation are applied to improve generalization. Following previous works \cite{BBCU,BRVE}, training is performed for 100K iterations using the Adam optimizer~\cite{Adam} with $\beta=0.9$, and a weight decay of $1\times10^{-4}$. A cosine annealing learning rate schedule \cite{cosineLR} is adopted, with the learning rate set to $2 \times 10^{-4}$ for all tasks. All experiments are conducted using PyTorch on two NVIDIA RTX 3090 GPUs.

Across the four RAW video restoration tasks, we use the same BinRVR backbone, bit-width setting, temporal window size, stride, optimizer, training iterations, learning rate schedule, patch size, batch size, and data augmentation strategy. The default temporal window size is 3 and the recurrent stride is 1. The only task-specific architectural adjustment is made for RAW video super-resolution, where a PixelShuffle upsampling layer is used in the reconstruction head to produce high-resolution RAW outputs.

\vspace{1mm}\noindent\textbf{Binarized comparison methods.}
For the four RAW video restoration tasks, we compare our BinRVR with state-of-the-art binarized methods, including BNN~\cite{BNN}, IRNet~\cite{IRNet}, Bireal~\cite{Bireal}, ReActNet~\cite{ReActNet}, BTM~\cite{BTM}, BBCU~\cite{BBCU}, BiSRNet~\cite{BiSRNet}, FRBC~\cite{FRBC}, BHViT~\cite{BHViT}, and BRVE~\cite{BRVE}, where BRVE is designed for low-light RAW video enhancement. As no existing BNN-based methods are specifically designed for RAW video restoration, we replace our Binarized Information Interaction Module (BIIM) and Distribution-Aware Binarized Convolution (DAB-Conv) in BinRVR with the corresponding binarized convolutional blocks of these methods, while keeping the remaining architecture unchanged to enable an isolated comparison of binarization strategies.

\vspace{1mm}\noindent\textbf{Evaluation metrics.} 
To evaluate the video restoration quality of each method, we calculate the average peak signal-to-noise ratio (PSNR), structural similarity index measure (SSIM), and spatio-temporal reduced reference entropic differences (ST-RRED)~\cite{STRRED} across all output RAW frames. The ST-RRED metric assesses video quality by capturing both spatial and temporal quality degradations. Higher values indicate better performance for PSNR and SSIM, while lower values indicate better performance for ST-RRED.

\vspace{1mm}\noindent\textbf{Model efficiency evaluation.} Following~\cite{BiSRNet, Bireal}, we estimate the model efficiency of binarized networks by weighting the binary operations with $1/64$ of the corresponding floating point operations (FLOPs) and the parameters (Params) with $1/32$ of their full-precision counterparts. For multi-bit quantization, the model efficiency is estimated by scaling the full-precision operations and parameters according to the bit width. Specifically, $2$-, $3$-, and $4$-bit operations are weighted by $1/16$, $3/32$, and $1/8$ of the full-precision FLOPs and parameters, respectively. The total FLOPs and parameters are obtained by summing the binarized (or quantized) and full-precision components. For all methods, per-frame FLOPs are computed by processing a $100$-frame RAW video with a spatial resolution of $256 \times 256$.

\subsection{Low-Light RAW Video Enhancement Results}
Low-light RAW video enhancement is a fundamental task in video restoration, aiming to recover clear, consistent content from sequences captured under insufficient illumination.

\vspace{1mm}\noindent\textbf{Datasets.}
We evaluate our BinRVR on two challenging low-light RAW video datasets, \textit{i.e.}, SMOID~\cite{SMOID} and LLRVD~\cite{LLRVD}. SMOID contains $309$ video pairs across $103$ indoor and outdoor scenes, each recorded at three ADC gain levels ($0$, $15$, and $30$) at $480 \times 640$ resolution. The dataset is divided into 70 scenes for training, $16$ for validation, and $17$ for testing. LLRVD consists of $210$ video pairs from $70$ high-resolution scenes at $1400\times2600$ resolution. For each scene, three low-light sequences are generated by fixing the ISO and varying the exposure time to simulate different illumination conditions. LLRVD is split into $60$ scenes for training, $4$ for validation, and $6$ for testing.

\vspace{1mm}\noindent\textbf{Quantitative results.}
We conduct a quantitative comparison of the two datasets for low-light RAW video enhancement, and the results are shown in Table~\ref{low-light-quantitative}. Our method consistently outperforms all binarized competitors across all settings and metrics. Compared with BRVE \cite{BRVE}, which is specifically designed for this task, our model achieves an average improvement of $0.18$~dB in PSNR. Compared with the recent binarized methods, BHViT \cite{BHViT}, our method yields an average PSNR gain of $0.40$~dB, clearly establishing its advantage in restoration quality. Moreover, our model attains these results with the lowest FLOPs and parameter count among all binarized methods, demonstrating its high efficiency.

Compared with full-precision low-light video enhancement methods, including SMOID~\cite{SMOID}, RViDeNet~\cite{RViDeNet}, FastDVD~\cite{FastDVD}, EMVD-L / EMVD-S~\cite{EMVD}, LLRVD~\cite{LLRVD}, FloRNN~\cite{FloRNN}, and ShiftNet~\cite{ShiftNet}, our method achieves a highly favorable accuracy-efficiency trade-off. In particular, when compared with the best full-precision model, ShiftNet~\cite{ShiftNet}, our method reduces FLOPs and parameters by over $96\%$ and $98\%$, respectively, while incurring only about a $4\%$ average PSNR drop. Moreover, compared with EMVD-S~\cite{EMVD}, a full-precision model with similar computational cost, our method attains up to $0.9$~dB higher PSNR together with consistent improvements in perceptual quality metrics.

\vspace{1mm}\noindent\textbf{Qualitative results.}
To further illustrate the visual quality achieved by different methods, Fig.~\ref{fig:vis_ll} presents qualitative comparisons on the SMOID \cite{SMOID} and LLRVD \cite{LLRVD} datasets. In the SMOID example, our method more effectively restores reflective objects within window regions, while most competing binarized methods fail to recover these reflections and suffer from severe detail degradation. This demonstrates that our method exhibits stronger enhancement in ultra-dark areas, producing clearer structures with fewer visual artifacts. On the LLRVD example, our method achieves more accurate color restoration along character boundaries, resulting in sharper edges and improved consistency with the ground truth. These observations provide clear visual evidence of the advantages of the proposed method under challenging low-light conditions.

\subsection{RAW Video Denoising Results}
RAW video denoising suppresses sensor noise while preserving details and temporal coherence. Compared with RGB data, RAW inputs retain linear sensor responses and richer scene information for restoration.

\vspace{1mm}\noindent\textbf{Datasets.}
Experiments are conducted on the CRVD~\cite{CRVD} dataset, which provides a challenging benchmark for RAW video denoising. CRVD contains $55$ groups of noisy-clean video pairs, each consisting of $7$ frames at a resolution of $1920 \times 1080$ and $20$ fps, captured using an IMX385 sensor. Noisy frames are recorded under high ISO settings ranging from $1600$ to $25600$, and the corresponding clean frames are obtained by averaging multiple noisy observations. The dataset includes $6$ indoor scenes for training and $5$ indoor scenes for validation and testing, providing diverse conditions for evaluating both spatial fidelity and temporal consistency.

\begin{table*}[t]
\centering
\footnotesize
\definecolor{lightgray}{gray}{0.94}
\setlength{\tabcolsep}{0.6mm}
\renewcommand{\arraystretch}{1.13}
\caption{Quantitative comparison on CRVD \cite{CRVD} for RAW video denoising. For binarized methods, the best and second-best results are shown in \textbf{bold} and \underline{underlined}, respectively. Lightweight full-precision methods are additionally included for reference, with the best result shown in \textbf{bold}. The results of full-precision methods are reported from the original papers.}
\vspace{-2mm}
\definecolor{lightgray}{gray}{0.94} % 定义浅灰色
\begin{NiceTabular}{c|c|ccccc|ccccccccc>{\columncolor{lightgray}}c}%四个c代表有四列且内容居中
\toprule
Method & Noisy & \makecell[c]{FastDVD \\ \cite{FastDVD}}  & \makecell[c]{EMVD-S \\ \cite{EMVD}} & \makecell[c]{RViDeNet \\ \cite{RViDeNet}} & \makecell[c]{LLRVD \\ \cite{LLRVD}} & \makecell[c]{FloRNN \\ \cite{FloRNN}} & \makecell[c]{BNN \\ \cite{BNN}} & \makecell[c]{Bireal \\ \cite{Bireal}} & \makecell[c]{IRNet \\ \cite{IRNet}} & \makecell[c]{ReActNet \\ \cite{ReActNet}} & \makecell[c]{BTM \\ \cite{BTM}} & \makecell[c]{BBCU \\ \cite{BBCU}} & \makecell[c]{BiSRNet \\ \cite{BiSRNet}} & \makecell[c]{FRBC \\ \cite{FRBC}} & \makecell[c]{BHViT \\ \cite{BHViT}} & \makecell[c]{\textbf{BinRVR} \\ \textbf{(Ours)}}\\
\midrule
Params(M)$\downarrow$ & - & 2.48 & \textbf{0.81} & 8.57 & 6.29 & 10.49 & 0.26 & 0.26 & 0.26 & 0.28 & \underline{0.25} & 0.27 & 0.28 & 0.26 & 0.27 & \textbf{0.22} \\
FLOPs(G)$\downarrow$ & - & 10.50 & \textbf{1.66} & 65.77 & 46.14 & 24.57 & 1.38 & 1.38 & 1.38 & 1.56 & \underline{1.33} & 1.47 & 1.44 & 1.38 & 
1.47 & \textbf{1.23} \\
\midrule
PSNR$\uparrow$ & 32.02 & 39.84 & 42.63 & 43.97 & 44.23 & \textbf{45.15} & 42.05 & 42.06 & 41.97 & 41.92 & 43.11 & \underline{43.21} & 43.16 & 41.48 & 43.17 & \textbf{43.53}\\
SSIM$\uparrow$ & 0.7333 & 0.9703 & 0.9851 & 0.9874 & 0.9879 & \textbf{0.9907} & 0.9804 & 0.9809 & 0.9800 & 0.9805 & 0.9847 & \underline{0.9850} & 0.9848 & 0.9779 & 0.9849 & \textbf{0.9868}\\
ST-RRED$\downarrow$  & 0.0171 & - & - & - & - & - & 0.0118 & 0.0109 & 0.0119 & 0.0116 & 0.0103 & \underline{0.0100} & \underline{0.0100} & 0.0119 & \underline{0.0100} & \textbf{0.0080} \\
\bottomrule
\end{NiceTabular}
\label{denoise-quantitative}
\end{table*}

\begin{figure*}[t]
    \centering
    \includegraphics[width=1.0\linewidth]{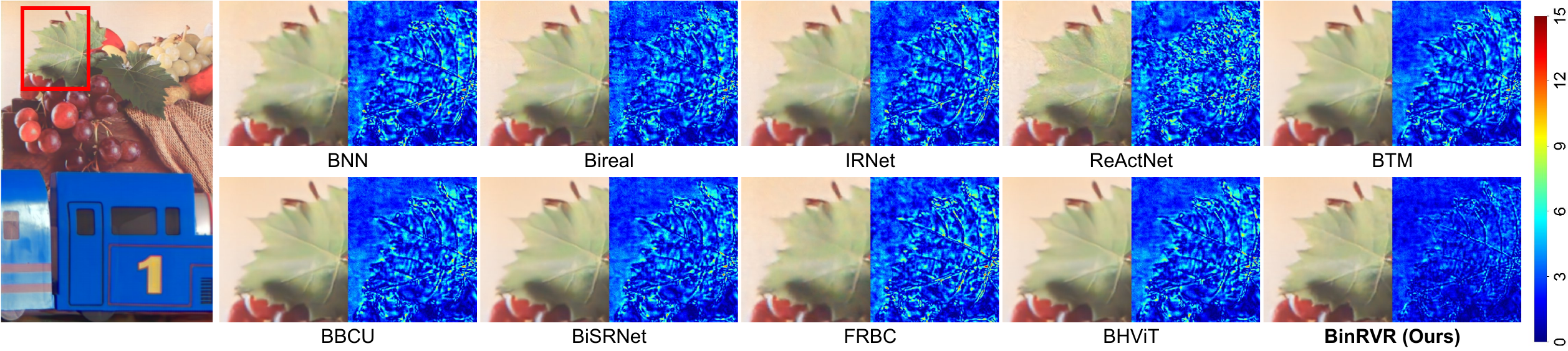}
    \vspace{-7mm}
    \caption{Visual comparisons of binarized methods on the CRVD~\cite{CRVD} dataset for RAW video denoising. For each example, the left half shows zoomed-in RGB results obtained by applying an ISP pipeline to the RAW outputs, while the right half presents the corresponding error maps with respect to the ground truth.}
    \label{fig:vis_denoise}
    \vspace{-2mm}
\end{figure*}

\vspace{1mm}\noindent\textbf{Quantitative results.}
We evaluate RAW video denoising performance on the CRVD \cite{CRVD}, with results summarized in Table~\ref{denoise-quantitative}. Our method outperforms all binarized methods across all metrics. Compared with the best-performing binarized competitor, BBCU \cite{BBCU}, it achieves a $0.32$~dB PSNR gain, higher SSIM, and a lower ST-RRED. We also compare with lightweight full-precision methods, including FastDVD \cite{FastDVD}, EMVD-S \cite{EMVD}, RViDeNet \cite{RViDeNet}, LLRVD \cite{LLRVD}, and FloRNN \cite{FloRNN}, whose results are taken from the original papers and therefore do not report ST-RRED. Compared with EMVD-S, which has a similar computational cost, our method attains $0.90$~dB higher PSNR. This indicates that our method achieves high efficiency while maintaining better performance.

\vspace{1mm}\noindent\textbf{Qualitative results.}
To qualitatively assess the denoising performance, Fig.~\ref{fig:vis_denoise} presents visual comparisons of RAW video denoising results. Our method produces noticeably cleaner results with finer and more clearly preserved leaf textures after denoising. The error maps exhibit a substantially larger proportion of blue regions, indicating lower reconstruction errors compared with other binarized methods. These visual differences highlight the clear advantage of our method in suppressing noise while preserving fine structural details.

\subsection{RAW Video Deblurring Results}
RAW video deblurring removes motion blur caused by camera or object motion to produce sharper and more stable videos. RAW inputs avoid ISP-induced artifacts and preserve a more faithful scene representation.

\vspace{1mm}\noindent\textbf{Datasets.}
Experiments are conducted on the Deblur-RAW~\cite{Deblur_RAW} dataset, collected across diverse indoor and outdoor scenes using a Canon EOS 6D camera. It consists of $103$ video sequences (around 100 frames per sequence), with $88$ sequences for training and $15$ for testing. Blurry inputs are produced by averaging several successive RAW frames, while the center frame of each sequence is used as the sharp reference. This construction preserves rich structural information and provides both RAW inputs and high-quality ground-truth targets, serving as a reliable benchmark for RAW video deblurring.

\begin{table*}[t]
\centering
\footnotesize
\definecolor{lightgray}{gray}{0.94}
\setlength{\tabcolsep}{1.3mm}
\renewcommand{\arraystretch}{1.13}
\caption{Quantitative comparison on Deblur-RAW \cite{Deblur_RAW} for RAW video deblurring. For binarized methods, the best and second-best results are shown in \textbf{bold} and \underline{underlined}, respectively. Lightweight full-precision methods are additionally included for reference, with the best result shown in \textbf{bold}.}
\vspace{-3mm}
\definecolor{lightgray}{gray}{0.94} % 定义浅灰色
\begin{NiceTabular}{c|c|ccc|ccccccccc>{\columncolor{lightgray}}c}%四个c代表有四列且内容居中
\toprule
Method & Blurry & \makecell[c]{MIMO \\ \cite{MIMO}} & \makecell[c]{ESTRNN \\ \cite{ESTRNN}} & \makecell[c]{LIEDNet \\ \cite{LIEDNet}} & \makecell[c]{BNN \\ \cite{BNN}} & \makecell[c]{Bireal \\ \cite{Bireal}} & \makecell[c]{IRNet \\ \cite{IRNet}} & \makecell[c]{ReActNet \\ \cite{ReActNet}} & \makecell[c]{BTM \\ \cite{BTM}} & \makecell[c]{BBCU \\ \cite{BBCU}} & \makecell[c]{BiSRNet \\ \cite{BiSRNet}} & \makecell[c]{FRBC \\ \cite{FRBC}} & \makecell[c]{BHViT \\ \cite{BHViT}} & \makecell[c]{\textbf{BinRVR} \\ \textbf{(Ours)}}\\
\midrule
Params(M) & - & 6.81 & \textbf{2.53} & 4.76 & 0.26 & 0.26 & 0.26 & 0.28 & \underline{0.25} & 0.27 & 0.28 & 0.26 & 0.27 & \textbf{0.22}\\
FLOPs(G) & - & 16.00 & 12.77 & \textbf{5.09} & 1.38 & 1.38 & 1.38 & 1.56 & \underline{1.33} & 1.47 & 1.44 & 1.38 & 1.47 & \textbf{1.23}\\
\midrule
PSNR$\uparrow$ & 36.52 & 38.15 & 38.12 & \textbf{38.61} & 36.88 & 36.81 & 36.95 & 37.33 & 37.35 & 37.50 & \underline{37.71} & 36.88 & 37.33 & \textbf{38.89}\\
SSIM$\uparrow$ & 0.9493 & 0.9663 & 0.9673 & \textbf{0.9708} &  0.9531 & 0.9520 & 0.9546 & 0.9583 & 0.9589 & 0.9607 & \underline{0.9626} & 0.9548 & 0.9587 & \textbf{0.9724}\\
ST-RRED$\downarrow$ & 0.0180 & 0.0151 & 0.0157 & \textbf{0.0098} & 0.0180 & \underline{0.0158} & 0.0179 & 0.0169 & 0.0179 & 0.0175 & 0.0167 & 0.0168 & 0.0169 & \textbf{0.0106} \\
\bottomrule
\end{NiceTabular}
\label{deblur-quantitative}
\vspace{-3mm}
\end{table*}

\vspace{1mm}\noindent\textbf{Quantitative results.}
We conduct RAW video deblurring experiments, and the results are summarized in Table~\ref{deblur-quantitative}. Our method delivers the best performance among all binarized methods, outperforming BBCU \cite{BBCU} by $1.18$~dB in PSNR. Consistent improvements are also observed in SSIM and ST-RRED. Compared to tasks such as low-light RAW video enhancement and RAW video denoising, the performance gap on this task is more pronounced, further demonstrating the advantages of our design under challenging motion blur. 

Moreover, we additionally compare against lightweight full-precision methods, including MIMO \cite{MIMO}, ESTRNN \cite{ESTRNN}, and LIEDNet \cite{LIEDNet}. Note that these methods are originally designed for RGB-domain deblurring and are adapted to the RAW-to-RAW setting due to the lack of lightweight full-precision deblurring methods specifically designed for RAW video restoration. Their lower performance does not imply an inherent advantage of binarization over full-precision models, but mainly reflects the domain gap between RGB deblurring and RAW video restoration.

In contrast, BinRVR is specifically designed for RAW video restoration, where BIIM models spatio-temporal information in the RAW domain and DAB-Conv mitigates quantization errors caused by activation distribution variations. Following Deblur-RAW \cite{Deblur_RAW}, we also adopt SSIM loss together with the reconstruction loss for deblurring, which encourages structural similarity and benefits blurred-detail recovery. Therefore, the superior result over these lightweight full-precision methods mainly comes from the RAW-oriented architecture and task-specific training strategy, while still maintaining substantially lower model size and FLOPs.

\begin{figure*}[t]
 \centering
 \includegraphics[width=1.0\linewidth]{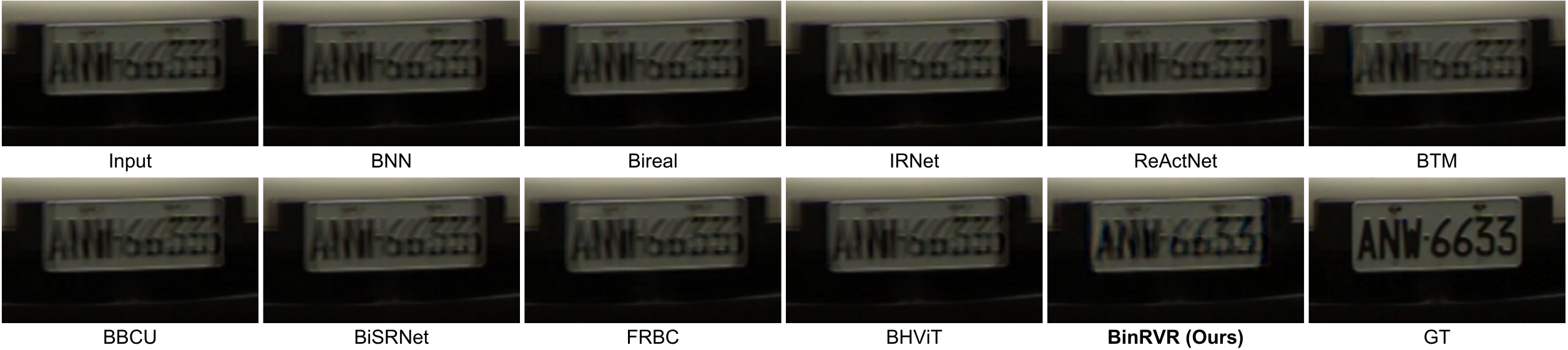}
 \vspace{-7mm}
 \caption{Visual comparisons of binarized methods on the Deblur-RAW~\cite{Deblur_RAW} dataset for RAW video deblurring.}
 \label{fig:vis_deblur}
 \vspace{-2mm}
\end{figure*}

\vspace{1mm}\noindent\textbf{Qualitative results.}
Visual comparisons in Fig.~\ref{fig:vis_deblur} demonstrate that our method recovers clearer alphanumeric details on the license plate, with results visually closer to the ground truth than other binarized methods. In comparison, binarized methods such as BiSRNet \cite{BiSRNet} and BHViT \cite{BHViT} introduce noticeable artifacts and exhibit blurred character edges.

\subsection{RAW Video Super-Resolution Results}
RAW video super-resolution reconstructs high-resolution sequences from low-resolution RAW inputs, requiring accurate spatial detail recovery and temporal coherence across frames.

\vspace{1mm}\noindent\textbf{Datasets.}
Experiments are conducted on the Real-RawVSR~\cite{Real-RawVSR} dataset, the first real-world RAW video super-resolution benchmark. It contains 450 LR-HR video pairs with $2\times$, $3\times$, and $4\times$ magnification. Each sequence includes about $150$ RAW frames captured using a dual-camera beam-splitter system, which avoids parallax and provides realistic degradations under diverse scenes and motions. The dataset also offers aligned sRGB counterparts, enabling comprehensive evaluation of RAW-domain super-resolution methods under real-world conditions.

\vspace{1mm}\noindent\textbf{Quantitative results.}
We evaluate RAW video super-resolution at $2\times$, $3\times$, and $4\times$ scales on Real-RawVSR~\cite{Real-RawVSR}, as shown in Table~\ref{sr-quantitative}. Among binarized methods, our method achieves the best performance across all metrics at $2\times$ and $3\times$ scales. At the more challenging $4\times$ scale, our method attains the highest PSNR and ST-RRED and ranks second-best in SSIM. Compared with the best-performing binarized alternatives at each scale (\textit{e.g.}, BHViT at $2\times$), our model improves PSNR by approximately $1.03$~dB at $2\times$ and $1.32$~dB at $3\times$, which exceeds the performance gaps typically observed among existing binarized solutions.

\begin{table*}[t]
\centering
\footnotesize
\setlength{\tabcolsep}{1.7mm}
\renewcommand{\arraystretch}{1.12}
\caption{Quantitative comparison on Real-RawVSR \cite{Real-RawVSR} for RAW video super-resolution. For binarized methods, the best and second-best results are shown in \textbf{bold} and \underline{underlined}, respectively. Lightweight full-precision methods are additionally included for reference, with the best result shown in \textbf{bold}. Params and FLOPs are computed under the $2\times$ scaling setting. $\dag$ indicates that Realviformer \cite{Realviformer} and FRBC \cite{FRBC} fail to train at certain scales.}
\vspace{-2mm}
\definecolor{lightgray}{gray}{0.94} % 定义浅灰色
\definecolor{lightblue}{RGB}{220, 240, 255}
\begin{NiceTabular}{l|c|c||ccc|ccc|ccc}%四个c代表有四列且内容居中
\toprule
\multirow{2}{*}{Method} & \multirow{2}{*}{\makecell[c]{Params\\(M)$\downarrow$}} & \multirow{2}{*}{\makecell[c]{FLOPs\\(G)$\downarrow$}} & \multicolumn{3}{c|}{$2 \times$} & \multicolumn{3}{c|}{$3 \times$} & \multicolumn{3}{c}{$4 \times$} \\

& & & PSNR$\uparrow$ & SSIM$\uparrow$ & ST-RRED$\downarrow$ & PSNR$\uparrow$ & SSIM$\uparrow$ & ST-RRED$\downarrow$ & PSNR$\uparrow$ & SSIM$\uparrow$ & ST-RRED$\downarrow$ \\
\midrule 
Bicubic & - & - & 35.23 & 0.9434 & 0.2702 & 33.40 & 0.9121 & 0.2827 & 31.25 & 0.8558 & 0.7312 \\
\midrule
\multicolumn{12}{l}{\textit{\textbf{Full-precision} methods}} \\
TDAN \cite{TDAN} & \textbf{1.86} & \textbf{13.76} & 34.19 & 0.9344 & 0.4757 & 33.06 & 0.9203 & 0.5424 & 32.13 & 0.8882 & 0.9604 \\
BasicVSR++ \cite{BasicVSR++} & 7.18 & 113.13 & 35.71 & 0.9579 & 0.3684 & \textbf{35.42} & \textbf{0.9345} & \textbf{0.3096} & 33.91 & 0.9121 & 0.5131 \\
Realviformer \cite{Realviformer} $^\dag$ & 5.82 & 40.46 & \textbf{37.53} & \textbf{0.9729} & \textbf{0.1720} & - & - & - & \textbf{34.57} & \textbf{0.9305} & \textbf{0.3654} \\
\midrule 
\multicolumn{12}{l}{\textit{\textbf{Binarized} methods}} \\
BNN \cite{BNN} & \underline{0.26} & 1.47 & 35.73 & 0.9537 & 0.2712 & 33.80 & 0.9282 & 0.3380 & 31.55 & 0.8941 & 0.7782 \\

Bireal \cite{Bireal} & \underline{0.26} & 1.47 & 33.98 & 0.9489 & 0.3867 & 34.08 & 0.9357 & \underline{0.3020} & 30.98 & 0.8973 & 0.6905 \\

IRNet \cite{IRNet} & \underline{0.26} & 1.47 & 35.42 & 0.9509 & 0.3161 & 33.48 & 0.9263 & 0.3333 & 32.01 & 0.8966 & 0.7658 \\

ReActNet \cite{ReActNet} & 0.29 & 1.65 & 35.93 & 0.9538 & 0.3175 & 34.38 & 0.9414 & 0.3094 & 32.39 & 0.9126 & 0.5722 \\

BTM \cite{BTM} & \underline{0.26} & \underline{1.43} & 33.94 & 0.9481 & 0.4886 & 34.24 & 0.9382 & 0.3993 & \underline{33.51} & \textbf{0.9258} & 0.5330 \\

BBCU \cite{BBCU} & 0.28 & 1.57 & 34.84 & 0.9515 & 0.4079 & 34.38 & \underline{0.9416} & 0.4305 & 33.04 & 0.9187 & 0.5919 \\

BiSRNet \cite{BiSRNet} & 0.28 & 1.53 & 34.82 & 0.9469 & 0.3942 & 33.82 & 0.9377 & 0.4310 & \underline{33.51} & 0.9224 & \underline{0.4695} \\

FRBC \cite{FRBC}$^\dag$ & \underline{0.26} & 1.47 & 35.67 & 0.9503 & 0.2464 & 32.60 & 0.8823 & 0.4607 & - & - & - \\

BHViT \cite{BHViT} & 0.28 & 1.56 & \underline{36.24} & \underline{0.9578} & \underline{0.2397} & \underline{34.85} & 0.9411 & 0.3534 & 33.38 & 0.9158 & 0.5780 \\

\rowcolor{lightgray}
\textbf{BinRVR (Ours)} & \textbf{0.23} & \textbf{1.33} & \textbf{37.27} & \textbf{0.9688} & \textbf{0.1882} & \textbf{36.17} & \textbf{0.9537} & \textbf{0.1788} & \textbf{34.27} & \underline{0.9232} & \textbf{0.3260} \\

\bottomrule
\end{NiceTabular}
\label{sr-quantitative}
\vspace{-3mm}
\end{table*}

We further compare against lightweight full-precision models, including TDAN~\cite{TDAN}, BasicVSR++~\cite{BasicVSR++}, and Realviformer~\cite{Realviformer}. Our method consistently outperforms TDAN and BasicVSR++ across all scales while maintaining significantly lower computational complexity. Compared to Realviformer, our method shows a modest performance drop but benefits from a substantially more compact design, with parameters about $25\times$ smaller. This demonstrates that our method delivers high efficiency and competitive performance consistently across restoration tasks.

\begin{figure*}[ht]
\centering
\includegraphics[width=1.0\linewidth]{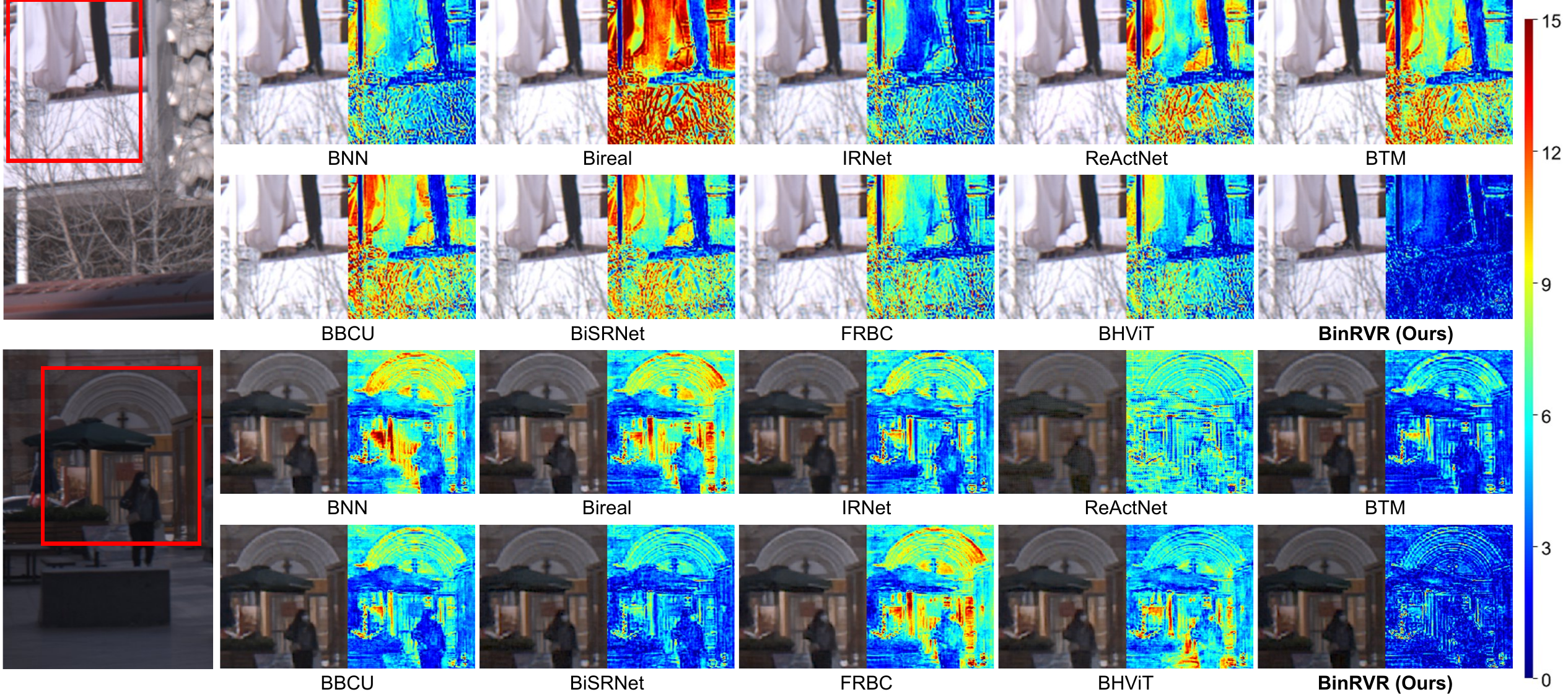}
 \vspace{-7mm}
 \caption{Visual comparisons of binarized methods on Real-RawVSR~\cite{Real-RawVSR} for RAW video super-resolution (top: $2\times$; bottom: $3\times$). For each example, the left half shows zoomed-in RGB results obtained by applying an ISP pipeline to the RAW outputs, while the right half presents the corresponding error maps with respect to the ground truth.}
 \label{fig:vis_sr}
 \vspace{-2mm}
\end{figure*}

\vspace{1mm}\noindent\textbf{Qualitative results.}
Qualitative comparisons are shown in Fig.~\ref{fig:vis_sr}. In the $2\times$ RAW video super-resolution case, our method reconstructs the fine textures of the white curtain more faithfully, while competing binarized methods tend to produce over-brightened results with missing shadow details and noticeable texture loss. For the $3\times$ case, our method restores the textures along the door frame with higher fidelity, exhibiting clearer structural patterns and more consistent shading. In contrast, BNN \cite{BNN}, Bireal \cite{Bireal}, and FRBC \cite{FRBC} results show incorrect smoothing on the same region, causing texture loss and muted boundaries, indicating inferior reconstruction of fine-grained details under higher magnification.

\subsection{Multi-Bit Quantization Experimental Results}
To further explore the adaptability of our framework to different deployment scenarios, we evaluate its multi-bit quantization capability using lightweight low-bit configurations (\textit{i.e.}, 2-bit, 3-bit, and 4-bit for both weights and activations). We compare our method against state-of-the-art quantization methods, including LSQ~\cite{LSQ}, QuantSR~\cite{QuantSR}, and Q-SCI~\cite{Q-SCI}, on the LLRVD~\cite{LLRVD} dataset.

As shown in Table~\ref{multi-bit-quantative}, our method consistently outperforms all counterparts across all bit-width settings. On average, our method improves PSNR by approximately $0.5$~dB over the best competing method, alongside noticeable gains in SSIM and ST-RRED. These results demonstrate that our quantization design remains effective beyond the binarized regime, preserving competitive restoration quality while retaining the advantages of low-bit efficiency.

\begin{table*}[t]
\centering
\footnotesize
\definecolor{lightgray}{gray}{0.94} % 定义浅灰色
\setlength{\tabcolsep}{1.5mm}
\renewcommand{\arraystretch}{1.12}
\caption{Quantitative results of downstream applications of low-light RAW video enhancement using the SMOID \cite{SMOID} dataset. We provide results for object detection and depth estimation. The best and second-best values are \textbf{Bold} and \underline{Underline}, respectively.}
\vspace{-2mm}
\begin{NiceTabular}{ll|ccccccccccc>{\columncolor{lightgray}}c}
\toprule
\textbf{Tasks} & \textbf{Metrics} & \makecell[c]{Linear \\ Scaled} & \makecell[c]{BNN \\ \cite{BNN}} & \makecell[c]{Bireal \\ \cite{Bireal}} & \makecell[c]{IRNet \\ \cite{IRNet}} & \makecell[c]{ReActNet \\ \cite{ReActNet}} & \makecell[c]{BTM \\ \cite{BTM}} & \makecell[c]{BBCU \\ \cite{BBCU}} & \makecell[c]{BiSRNet \\ \cite{BiSRNet}} & \makecell[c]{FRBC \\ \cite{FRBC}} & \makecell[c]{BHViT \\ \cite{BHViT}} & \makecell[c]{BRVE \\ \cite{BRVE}} & \makecell[c]{\textbf{BinRVR} \\ \textbf{(Ours)}}\\
\midrule
\multirow{3}{*}{Object Detection} 
& AP$\uparrow$   & 61.98 & 62.85 & 65.02 & 63.49 & 65.46 & 66.83 & 66.61 & 66.60 & 53.39 & 66.97 & \underline{68.52} & \textbf{70.62} \\
& AP$_{50}$$\uparrow$   & 64.93 & 65.63 & 67.91 & 66.29 & 68.15 & 69.77 & 69.44 & 69.59 & 56.31 & 69.84 & \underline{71.05} & \textbf{73.24} \\
& AP$_{75}$$\uparrow$   & 63.34 & 64.22 & 66.67 & 64.98 & 66.96 & 68.49 & 68.27 & 68.27 & 54.67 & 68.66 & \underline{69.97} & \textbf{72.11} \\
\midrule
\multirow{2}{*}{Depth Estimation}
& RMSE$\downarrow$   & 0.1308 & 0.1251 & 0.1270 & 0.1255 & 0.1224 & 0.1136 & 0.1115 & 0.1149 & 0.2215 & 0.1127 & \underline{0.1067} & \textbf{0.1029} \\
& SSIM$\uparrow$   & 0.9324 & 0.9335 & 0.9313 & 0.9323 & 0.9342 & 0.9382 & 0.9392 & 0.9372 & 0.8770 & 0.9386 & \underline{0.9463} & \textbf{0.9482} \\
\bottomrule
\end{NiceTabular}
\label{tab:downstream}
\vspace{-4mm}
\end{table*}

\begin{table}[t]
\begin{center}
\caption{Multi-Bit quantization results on LLRVD \cite{LLRVD} dataset for low-light RAW video enhancement.}
\vspace{-2mm}
\label{multi-bit-quantative}
\footnotesize
\definecolor{lightgray}{gray}{0.94} % 定义浅灰色
\setlength{\tabcolsep}{0.7mm}
\renewcommand{\arraystretch}{1.15}
\begin{NiceTabular}{lc|cc|ccc}
\toprule
Method & Bits (w/a) & Params$\downarrow$ & FLOPs$\downarrow$  & PSNR$\uparrow$& SSIM$\uparrow$& ST-RRED$\downarrow$ \\ \midrule
LSQ \cite{LSQ}       & 4/4 & 0.80M & 5.81G & 37.10 & 0.9585 & 0.0461 \\
QuantSR \cite{QuantSR}  & 4/4 & 0.80M & 5.81G & 37.09 & 0.9585 & 0.0460 \\
Q-SCI \cite{Q-SCI}  & 4/4 & 0.80M & 5.81G & 37.13 & 0.9590 & 0.0433 \\
\rowcolor{lightgray} \textbf{Ours}       & 4/4   & 0.80M & 5.81G & \textbf{37.71} & \textbf{0.9646} & \textbf{0.0382} \\ \midrule
LSQ \cite{LSQ}       & 3/3 & 0.63M & 3.39G & 37.05 & 0.9580 & 0.0465 \\
QuantSR \cite{QuantSR} & 3/3  & 0.63M & 3.39G & 37.06 & 0.9581 & 0.0464  \\
Q-SCI \cite{Q-SCI}  & 3/3  & 0.63M & 3.39G & 37.10 & 0.9588 & 0.0460 \\
\rowcolor{lightgray} \textbf{Ours}  & 3/3 & 0.63M & 3.39G & \textbf{37.64} & \textbf{0.9641} & \textbf{0.0390} \\ \midrule
LSQ \cite{LSQ}       & 2/2 & 0.45M & 2.17G & 36.99 & 0.9573 & 0.0480 \\
QuantSR \cite{QuantSR} & 2/2 & 0.45M & 2.17G & 36.99  & 0.9573 & 0.0479 \\
Q-SCI \cite{Q-SCI}  & 2/2  & 0.45M & 2.17G & 37.01 & 0.9576 & 0.0475 \\
\rowcolor{lightgray} \textbf{Ours}& 2/2 & 0.45M & 2.17G & \textbf{37.53} & \textbf{0.9631} & \textbf{0.0403} \\ 
\bottomrule
\end{NiceTabular}
\vspace{-5mm}
\end{center}
\end{table}

% \vspace{-1mm}
\subsection{Downstream Video Applications}
To assess the practical value of our restoration framework, we examine how restored videos benefit representative downstream high-level video tasks. We consider object detection and monocular depth estimation, validating that low-light RAW video enhancement yields measurable gains in downstream performance. Our method is compared with state-of-the-art binarized methods under identical settings. Experiments are conducted on the large-scale SMOID \cite{SMOID} dataset to reduce statistical bias. For all tasks, restored RAW videos are first converted to RGB via ISP, matching typical usage in downstream video applications.

\begin{figure*}[t]
    \centering
    \includegraphics[width=1.0\textwidth]{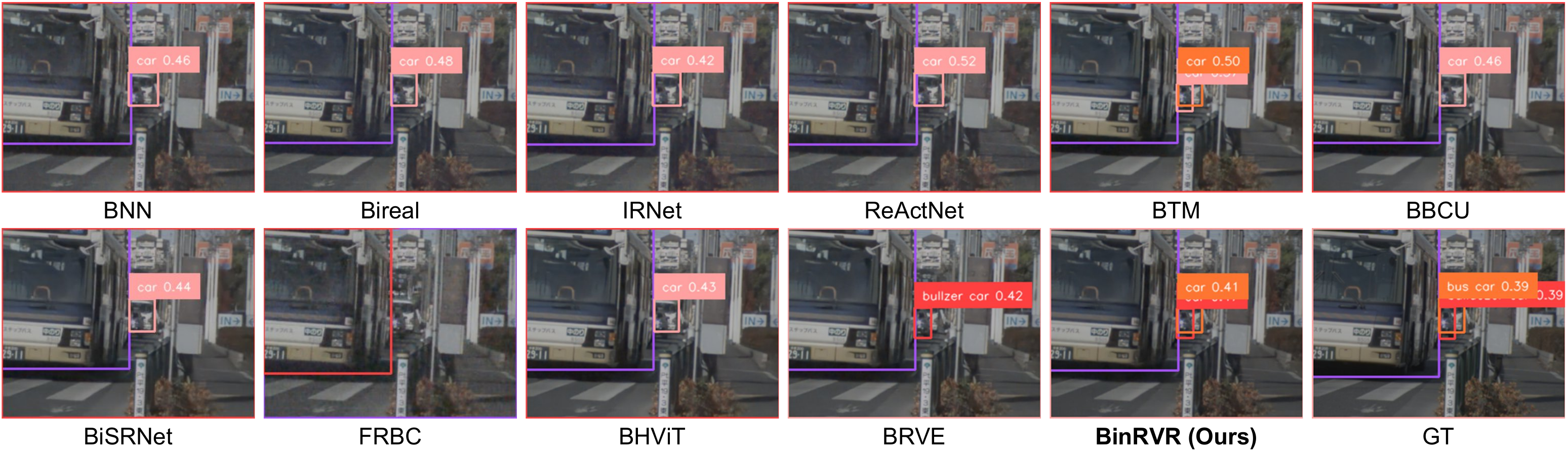}
    \vspace{-7mm}
    \caption{Qualitative comparisons of object detection results produced by GroundingDINO \cite{GroundingDINO} on enhanced videos.}
    \label{fig:detection}
    \vspace{-2mm}
\end{figure*}

\vspace{1mm}\noindent\textbf{Object detection.}
Object detection is a fundamental component in video understanding pipelines, providing essential semantic cues for higher-level tasks such as tracking and scene analysis. We employ GroundingDINO \cite{GroundingDINO}, a popular text-guided detector with strong zero-shot generalization. As a comprehensive prompt set, $1203$ categories from the LVIS \cite{LVIS} dataset are used as textual inputs. We use the SMOID \cite{SMOID} dataset’s normal-light videos and process them with GroundingDINO to obtain detection labels as reliable pseudo ground truth. Following previous works in object detection \cite{DETR}, performance is measured using AP, AP${50}$, and AP${75}$. As shown in the upper half of Table~\ref{tab:downstream}, our method achieves the highest scores across all metrics, outperforming all binarized competitors by a clear and consistent margin. Qualitative visualizations are provided in Fig.~\ref{fig:detection}, showing that our BinRVR yields more reliable detections for small, partially occluded targets in challenging scenes.

\begin{figure*}[t]
    \centering
    \includegraphics[width=1.0\textwidth]{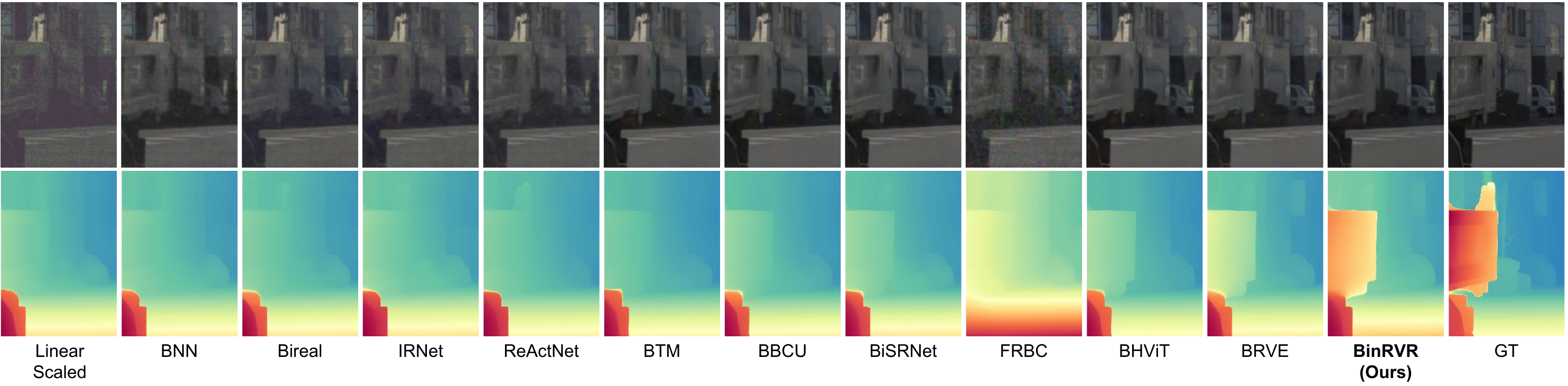}
    \vspace{-7mm}
    \caption{Qualitative comparisons of monocular depth estimation results produced by Depth Anything V2 \cite{DepthAnythingv2} on enhanced videos.}
    \vspace{-4mm}
    \label{fig:depth}
\end{figure*}

\vspace{1mm}\noindent\textbf{Monocular depth estimation.}
Monocular depth estimation provides crucial geometric cues for downstream applications such as 3D reconstruction and scene understanding. We adopt Depth Anything V2 \cite{DepthAnythingv2}, a widely used framework with strong zero-shot generalization, to evaluate how enhanced videos affect depth prediction accuracy. Ground-truth reference is generated by applying Depth Anything V2 to the SMOID \cite{SMOID} dataset’s normal-light videos. For quantitative evaluation, we calculate the root-mean-square error (RMSE) to measure numerical accuracy and structural similarity (SSIM) to assess structural consistency in the predicted depth maps. As reported in the lower half of Table~\ref{tab:downstream}, our method again surpasses all binarized counterparts, demonstrating more accurate geometric estimation and better structural fidelity. Qualitative visualizations are provided in Fig.~\ref{fig:depth}, where our method estimates depth more accurately on the rear part of the truck.

\subsection{Discussion}
We discuss several key aspects of our framework, including the effectiveness of RAW-domain restoration, temporal modeling design choices, computational overhead, and ablation studies on our BIIM and DAB-Conv.

\vspace{1mm}\noindent\textbf{Effectiveness of RAW-Domain restoration.}
We evaluate different signal representations for low-light video enhancement on the LLRVD~\cite{LLRVD} dataset, with results summarized in Table~\ref{tab:raw_setting}. Among all settings, RAW2RAW+ISP achieves the best performance in terms of all metrics. Preserving RAW signals as both the input and output of the restoration stage minimizes information loss introduced by early ISP processing and enables more accurate recovery. In contrast, RAW2RGB suffers from the joint learning of enhancement and color mapping, while RGB2RGB performs the worst due to irreversible quantization in low-intensity regions. When compression is applied, RAW-domain restoration remains robust, whereas restoring already compressed RGB videos leads to severe degradation. These results demonstrate that RAW-domain restoration provides the most reliable outcomes.

\begin{figure}[t]
    \centering
    \includegraphics[width=\columnwidth]{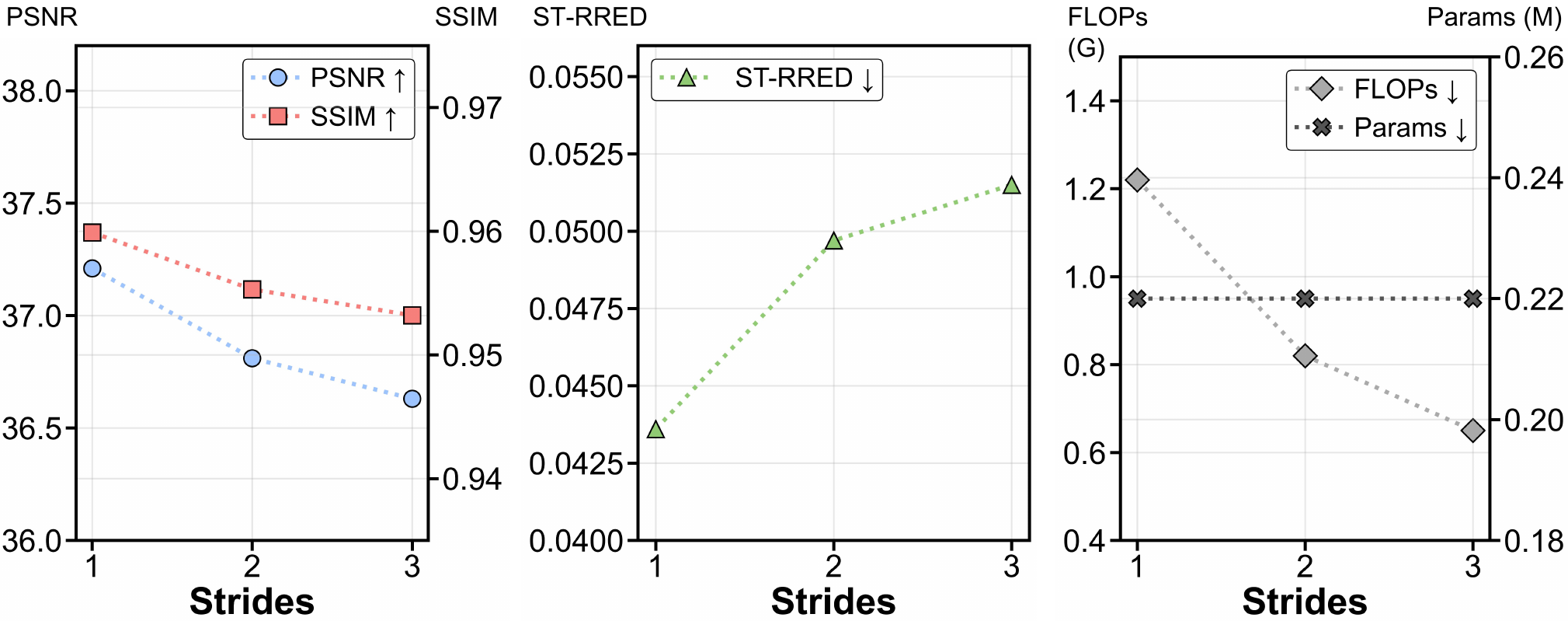}
    \vspace{-7mm}
    \caption{Effect of the sliding window stride on recurrent propagation. A larger stride improves efficiency but compromises temporal consistency in restored videos across frames.}
    \vspace{-2mm}
    \label{stride}
\end{figure}

\begin{table}[t]
\centering
\caption{Comparison of different low-light video enhancement settings on LLRVD \cite{LLRVD} dataset. Metrics are computed in the RGB domain after ISP processing.}
\vspace{-2mm}
\label{tab:raw_setting}
\footnotesize
\setlength{\tabcolsep}{3.0mm}
\renewcommand{\arraystretch}{1.17}
\begin{tabular}{lccc}
\toprule
Method & PSNR$\uparrow$ & SSIM$\uparrow$ & ST-RRED$\downarrow$ \\
\midrule
RAW2RAW+ISP       & \textbf{30.59} & \textbf{0.8408} & \textbf{0.2002} \\
RAW2RGB           & 27.53 & 0.8143 & 0.2938 \\
RGB2RGB           & 24.91 & 0.7994 & 0.5941 \\
\midrule
RAW2RAW+ISP+H264  & \textbf{30.24} & \textbf{0.8421} & \textbf{0.1810} \\
RGB2RGB+H264      & 24.88 & 0.8170 & 0.5610 \\
H264+RGB2RGB      & 20.15 & 0.6138 & 1.4199 \\
\bottomrule
\end{tabular}
\vspace{-2mm}
\end{table}

\vspace{1mm}\noindent\textbf{Effectiveness of sliding-window recurrent connection.}
To directly compare temporal modeling paradigms, we conduct an architecture-level ablation study under the same backbone, 1-bit quantization setting, and training configuration. As shown in Table~\ref{tab:temporal_arch_ablation}, the pure sliding-window structure is efficient but lacks long-term recurrent propagation, while the pure unidirectional recurrent structure improves temporal consistency but is weaker in short-term neighboring-frame aggregation. The proposed hybrid structure achieves the best PSNR, SSIM, and ST-RRED, demonstrating a better balance between short-term multi-frame aggregation and long-term temporal modeling.

\begin{table}[t]
\begin{center}
\caption{Architecture-level ablation study on the temporal modeling paradigm. All variants use the same backbone, 1-bit quantization setting, and training configuration.}
\vspace{-2mm}
\label{tab:temporal_arch_ablation}
\scriptsize
\definecolor{lightgray}{gray}{0.94}
\setlength{\tabcolsep}{3pt}
\renewcommand{\arraystretch}{1.12}
\begin{NiceTabular}{l|cc|cc|ccc}
\toprule
Arch. & Window & Rec. & Params$\downarrow$ & FLOPs$\downarrow$ & PSNR$\uparrow$ & SSIM$\uparrow$ & ST-RRED$\downarrow$ \\
\midrule
\makecell[l]{Pure sliding\\window} & 3 & \XSolidBrush & 0.22M & 0.74G & 36.83 & 0.9567 & 0.0648 \\
\makecell[l]{Pure recurrent} & 1 & \Checkmark & 0.22M & 1.05G & 36.98 & 0.9579 & 0.0506 \\
\rowcolor{lightgray} Hybrid (Ours) & 3 & \Checkmark & 0.22M & 1.23G & \textbf{37.21} & \textbf{0.9599} & \textbf{0.0436} \\
\bottomrule
\end{NiceTabular}
\end{center}
\vspace{-4mm}
\end{table}

\vspace{1mm}\noindent\textbf{Effect of stride in sliding window recurrent connection.}
In our BinRVR, the sliding-window recurrent connection of the BIIM decoder aggregates temporal cues across frames, and its behavior is largely determined by the stride. As shown in Fig.~\ref{stride}, a stride of $1$ maintains full recurrent propagation and delivers the best temporal fidelity. Increasing the stride to $2$ retains only the most recent feature for recurrence, reducing nearly $40\%$ FLOPs with a $0.40$~dB drop in PSNR, suggesting that recurrent updates are beneficial but not required at every step. With a stride $3$, recurrent links are removed entirely and each window is processed independently, yielding the highest efficiency but causing a sharp rise in ST-RRED due to temporal inconsistency. These observations confirm the effectiveness of the sliding window recurrent connection in modeling temporal information across frames.

\begin{table}[t]
\begin{center}
\caption{Ablation studies on our Binarized Information Interaction (BIIM) Module .}
\vspace{-2mm}
\label{ablation-este}
\footnotesize
\definecolor{lightgray}{gray}{0.94} % 定义浅灰色
\setlength{\tabcolsep}{0.1mm}
\renewcommand{\arraystretch}{1.13}
\begin{NiceTabular}{c|c|ccc|cc}
\toprule
Partition & Ablation & PSNR$\uparrow$ & SSIM$\uparrow$ & ST-RRED$\downarrow$ & Params$\downarrow$ & FLOPs$\downarrow$   \\ \midrule
\multirow{2}{*}{\makecell[c]{BIIM \\ Encoder}} & w/o TII & 37.06 & 0.9585 & 0.0468 & 0.22M & 1.23G   \\
 & Same as Decoder-TII & 37.15 & 0.9586 & 0.0450 & 0.24M & 1.43G \\ \midrule
\multirow{7}{*}{\makecell[c]{BIIM \\ Decoder}} & w/o TII & 37.00 & 0.9581 & 0.0470 & 0.21M & 1.12G \\
 & Same as Encoder-TII & 37.14 & 0.9595 & 0.0451 & 0.21M &  1.12G  \\ 
 & w/o SII & 37.04 & 0.9580 & 0.0469 & 0.21M & 1.17G  \\ 
 & Strip Kernel $\to 1\times9$  & 37.16 & 0.9592 & 0.0453 & 0.22M & 1.23G  \\ 
 & Strip Kernel $\to 1\times13$  & 37.17 & 0.9596 & \textbf{0.0426} & 0.22M & 1.23G \\ 
 & Branch Ratio $\to 1/4$  & 37.17 & 0.9594 & 0.0443 & 0.23M & 1.23G \\ 
 & Branch Ratio $\to 1/16$  & 37.15 & 0.9596 & 0.0447 & 0.22M & 1.23G  \\ 
  \midrule
 \rowcolor{lightgray}- & BinRVR (Ours)  & \textbf{37.21} & \textbf{0.9599} & 0.0436 & 0.22M & 1.23G  \\ 
\bottomrule
\end{NiceTabular}
\end{center}
\vspace{-3mm}
\end{table}

\vspace{1mm}\noindent\textbf{Long-sequence temporal consistency.}
To examine temporal consistency on long sequences, we evaluate BinRVR with different input lengths on the SMOID test set \cite{SMOID}. For fair comparison, all settings use the same 30-frame target interval for evaluation. Specifically, for an input length $L$, the model takes the segment ending at the same frame, i.e., frames $[t-L+1,\ldots,t]$, while PSNR, SSIM, and ST-RRED are computed only on the identical target interval. Thus, different lengths only change the amount of historical context provided to the recurrent connection. As shown in Table~\ref{tab:long_sequence}, the average PSNR and SSIM over three gain settings remain nearly stable when increasing the sequence length from $30$ to $120$ frames, and ST-RRED shows no sharp degradation. This indicates that BinRVR does not suffer from severe error accumulation within the tested range, benefiting from the proposed sliding-window recurrent connection that injects current local-window features at each step rather than relying solely on the propagated hidden state. Nevertheless, extremely long sequences with severe motion may still challenge the current unidirectional design and will be explored in future work.

\begin{table}[t]
\begin{center}
\caption{Long-sequence evaluation on the SMOID \cite{SMOID} test set. The metrics are averaged over Gain0, Gain15, and Gain30, and computed on the same target interval.}
\vspace{-2mm}
\label{tab:long_sequence}
\footnotesize
\setlength{\tabcolsep}{4.5mm}
\renewcommand{\arraystretch}{1.15}
\begin{tabular}{c|ccc}
\toprule
\multirow{2}{*}{\makecell[c]{Sequence\\Length}}
& \multicolumn{3}{c}{Avg. over Gain0 / Gain15 / Gain30} \\
\cline{2-4}
& PSNR$\uparrow$ & SSIM$\uparrow$ & ST-RRED$\downarrow$ \\
\midrule
30 frames  & 40.86 & 0.9789 & 0.0610 \\
60 frames  & 40.88 & 0.9790 & 0.0606 \\
90 frames  & 40.88 & 0.9790 & 0.0606 \\
120 frames & 40.88 & 0.9790 & 0.0605 \\
\bottomrule
\end{tabular}
\end{center}
\vspace{-4mm}
\end{table}

\begin{figure}[t]
\centering
\includegraphics[width=\columnwidth]{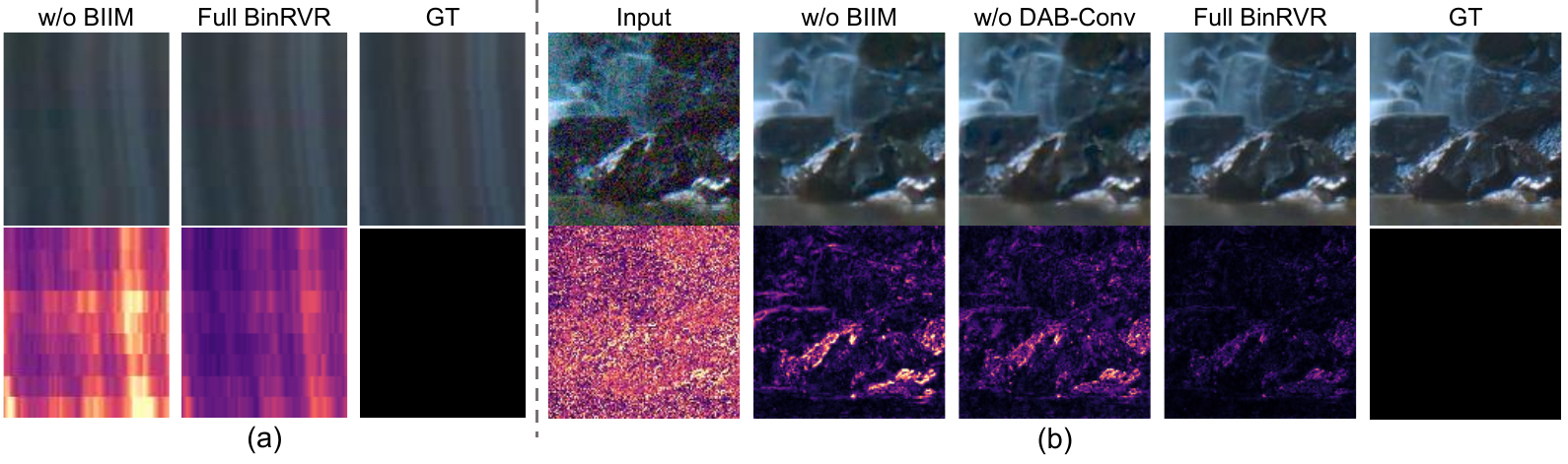}
\vspace{-8mm}
\caption{Visual ablation comparison of representative BinRVR variants. (a) Ablation of temporal modeling components. (b) Ablation of spatial modeling components. Each comparison includes the degraded input, ablation variants without key components, the full model, the ground truth, and the corresponding error maps.}
\label{fig:vis_ablation}
\vspace{-3mm}
\end{figure}

\vspace{1mm}\noindent\textbf{Effectiveness of binarized information interaction module.}
Table~\ref{ablation-este} reports ablations on BIIM. Removing Temporal Information Interaction (TII) in the decoder results in a PSNR drop of $0.21$~dB, which is notably larger than the $0.15$~dB drop caused by removing it in the encoder, indicating that temporal interaction in the decoder plays a more critical role in reconstruction. Eliminating Spatial Information Interaction (SII) reduces performance by $0.17$~dB, showing that spatial interaction remains beneficial under binarization. Using identical TII structures or modifying strip kernel sizes and branch ratios leads to only marginal changes while increasing parameters or FLOPs, indicating that the adopted asymmetric design and hyperparameter settings are already well balanced. The visual ablation results in Fig.~\ref{fig:vis_ablation} further show that removing BIIM leads to less accurate restoration and larger reconstruction errors, while the full BinRVR recovers cleaner structures closer to the ground truth. These results verify that all components of BIIM contribute positively to its effectiveness.

\begin{figure}[t]
\centering
\includegraphics[width=\columnwidth]{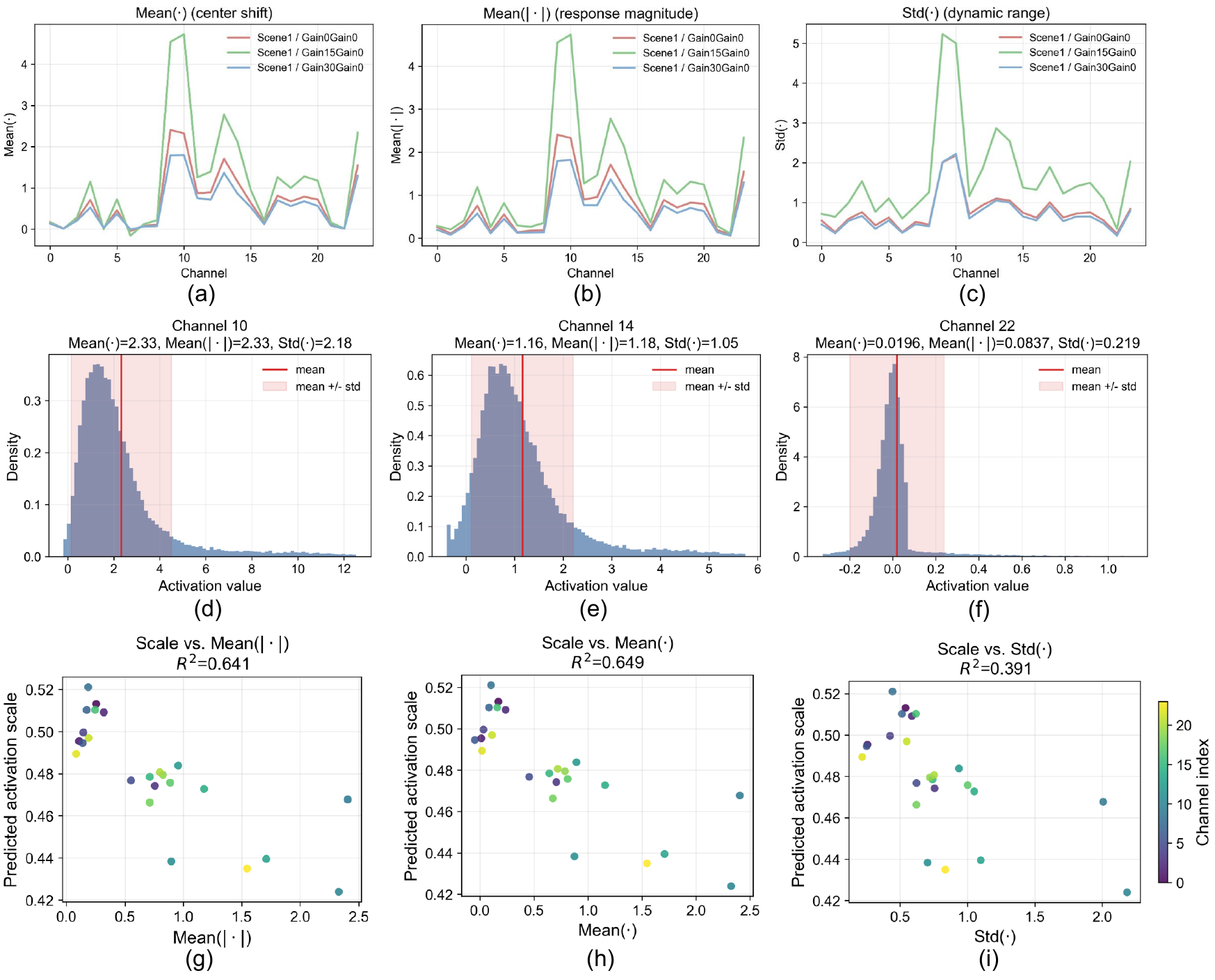}
\vspace{-8mm}
\caption{Visualization of activation distributions and DAB-Conv scaling factors. (a)--(c) Channel-wise activation statistics before binarization under different gain settings. (d)--(f) Representative activation distributions from different channels. (g)--(i) Predicted DAB-Conv scales versus activation statistics. The results show that activation distributions vary across channels and degradations, motivating input-dependent channel-wise scaling.}
\label{fig:dab_statistics}
\vspace{-3mm}
\end{figure}

\vspace{1mm}\noindent\textbf{Effectiveness of distribution-aware convolution.}
Table~\ref{ablation-dabc} evaluates the impact of different statistical descriptors in our Distribution-Aware Binarized Convolution. To further explain the motivation, Fig.~14 visualizes the activation statistics of a representative DAB-Conv layer and their relationship with the predicted activation scaling factors. The channel-wise mean, absolute mean, and standard deviation vary noticeably across channels and gain settings, while representative channel histograms show shifted, asymmetric, or narrowly concentrated activation distributions. These observations indicate that RAW video activations cannot be well approximated by a fixed or globally shared scale, and motivate the joint modeling of center shift, response magnitude, and dynamic range. Moreover, Fig.~\ref{fig:dab_statistics} shows non-trivial correlations between these statistics and the predicted scaling factors, suggesting that DAB-Conv learns input-dependent channel-wise modulation rather than applying a fixed scaling rule. The visual ablation in Fig.~\ref{fig:vis_ablation}\textcolor{blue}{(b)} further shows that removing DAB-Conv leads to less faithful restoration and larger reconstruction errors.

Consistent with this analysis, removing all statistics causes a PSNR drop of $0.17$~dB, indicating that distribution cues are necessary for stable binarization. Introducing only the mean term yields a $0.08$~dB improvement, and combining the mean with standard deviation further improves performance to a total gain of $0.13$~dB, confirming that joint modeling of first- and second-order statistics is beneficial. Incorporating $\text{Mean}(|\cdot|)$ provides additional gains and leads to the best performance. These results demonstrate that leveraging multiple statistical descriptors enables more reliable distribution-aware scaling.

\begin{table}[t]
\begin{center}
\caption{Ablation studies for our Distribution-Aware Binarized Convolution.}
\vspace{-2mm}
\label{ablation-dabc}
\scriptsize
\definecolor{lightgray}{gray}{0.94} % 定义浅灰色
\setlength{\tabcolsep}{0.6mm}
\renewcommand{\arraystretch}{1.13}
\begin{NiceTabular}{ccc|ccc|cc}
\toprule
$\text{Mean}(\cdot)$ & $\text{Std}(\cdot)$ & $\text{Mean}(|\cdot|)$ & PSNR$\uparrow$ & SSIM$\uparrow$ & ST-RRED$\downarrow$ & Params$\downarrow$ & FLOPs$\downarrow$ \\ \midrule

\XSolidBrush & \XSolidBrush & \XSolidBrush & 37.04 & 0.9581 & 0.0458 & 0.222M & 1.21G \\
\Checkmark & \XSolidBrush & \XSolidBrush & 37.12 & 0.9589 & 0.0447 & 0.222M & 1.21G \\
\Checkmark & \Checkmark & \XSolidBrush & 37.17 & 0.9595 & 0.0444 & 0.223M & 1.21G  \\
\midrule
\rowcolor{lightgray} \Checkmark & \Checkmark & \Checkmark & \textbf{37.21} & \textbf{0.9599} & \textbf{0.0436} & 0.223M & 1.23G  \\
\bottomrule
\end{NiceTabular}
\end{center}
\vspace{-2mm}
\end{table}

\vspace{1mm}\noindent\textbf{Deployment feasibility evaluation.}
Native end-to-end BNN execution is still rarely supported by mainstream deployment hardware and runtimes, because it typically requires dedicated bit-packing and XNOR-popcount kernels. Therefore, instead of reporting only normalized FLOPs and parameter counts, we further evaluate actual-runtime deployment on representative and accessible GPU, CPU, and mobile NPU platforms under currently supported inference settings, including FP32, W8A8, and W4A8. As summarized in Table~\ref{tab:deployment_results}, BinRVR consistently achieves higher throughput than the representative full-precision video restoration model across diverse hardware. Under FP32 inference, BinRVR improves FPS by about 24.9$\times$ on the RTX 4060 GPU, 15.1$\times$ on Jetson Orin NX, 20.9$\times$ on the Core Ultra 9 CPU, 10.9$\times$ on the Snapdragon 8 Elite CPU, 2.1$\times$ on the Snapdragon 8 Elite NPU, and 13.0$\times$ on the Ascend 310B4 NPU. Moreover, the W8A8 and W4A8 settings further improve actual throughput, with up to 4.2$\times$ FPS improvement over ShiftNet on the Snapdragon 8 Elite NPU. These results demonstrate that our BinRVR can be effectively mapped to practical low-bit inference paths under existing hardware and runtime constraints. Native 1-bit BNN acceleration may provide additional gains when dedicated kernels become more widely available.

\begin{table}[t]
\begin{center}
\caption{Actual-runtime deployment comparison with the representative full-precision method ShiftNet \cite{ShiftNet} on different hardware platforms under currently supported inference modes. PSNR is evaluated on the LLRVD \cite{LLRVD} test set.}
\vspace{-2mm}
\label{tab:deployment_results}
\scriptsize
\definecolor{lightgray}{gray}{0.94}
\setlength{\tabcolsep}{1.6mm}
\renewcommand{\arraystretch}{1.15}
\begin{tabular}{l|l|c|c|c|c}
\hline
Platform & Model & Mode & FPS$\uparrow$ & \shortstack[c]{\makecell[c]{Latency\\(ms/f.)$\downarrow$}} & PSNR$\uparrow$ \\
\hline
\multirow{2}{*}{\shortstack[l]{NVIDIA RTX 4060 GPU\\Local laptop}}
& ShiftNet & FP32 & 3.74 & 267.34 & 37.87 \\
& \textbf{Ours} & FP32 & 93.22 & 10.73 & 37.81 \\
\hline
\multirow{3}{*}{\shortstack[l]{NVIDIA Ampere GPU\\NVIDIA Jetson Orin NX}}
& ShiftNet & FP32 & 1.30 & 770.39 & 37.87 \\
& \textbf{Ours} & FP32 & 19.94 & 50.16 & 37.81 \\
& \textbf{Ours} & W8A8 & 35.18 & 28.42 & 37.77 \\
\hline
\multirow{3}{*}{\shortstack[l]{Intel Core Ultra 9 CPU\\Local laptop}}
& ShiftNet & FP32 & 0.87 & 1144.17 & 37.87 \\
& \textbf{Ours} & FP32 & 18.14 & 55.12 & 37.81 \\
& \textbf{Ours} & W8A8 & 20.96 & 47.70 & 37.77 \\
\hline
\multirow{3}{*}{\shortstack[l]{Snapdragon 8 Elite CPU\\Xiaomi 15}}
& ShiftNet & FP32 & 0.36 & 2793.80 & 37.81 \\
& \textbf{Ours} & FP32 & 3.94 & 253.58 & 37.79 \\
& \textbf{Ours} & W8A8 & 9.93 & 100.72 & 37.68 \\
\hline
\multirow{4}{*}{\shortstack[l]{Snapdragon 8 Elite NPU\\Qualcomm AI Hub}}
& ShiftNet & FP32 & 45.54 & 21.96 & 37.87 \\
& \textbf{Ours} & FP32 & 95.26 & 10.48 & 37.81 \\
& \textbf{Ours} & W8A8 & 154.85 & 6.46 & 37.77 \\
& \textbf{Ours} & W4A8 & 192.33 & 5.20 & 37.75 \\
\hline
\multirow{3}{*}{\shortstack[l]{Huawei Ascend 310B4 NPU\\Huawei Atlas 200I DK A2}}
& ShiftNet & FP32 & 2.62 & 382.25 & 37.87 \\
& \textbf{Ours} & FP32 & 34.18 & 29.26 & 37.81 \\
& \textbf{Ours} & W8A8 & 65.76 & 15.21 & 37.77 \\
\hline
\end{tabular}
\end{center}
\vspace{-4mm}
\end{table}

\section{Conclusion} \label{conclusion}
This paper presents BinRVR, a systematic high-efficiency binarized framework for multi-task RAW video restoration that explicitly targets the unique challenges of temporal modeling and distribution mismatch under extreme quantization. Unlike existing binarized networks that mainly focus on image-level tasks, BinRVR is specifically designed for video restoration in the RAW domain, where preserving temporal coherence and fine-grained sensor information is critical. To this end, we introduce a Binarized Information Interaction Module (BIIM) to enable efficient joint modeling of inter-frame temporal dependencies and intra-frame spatial structures using lightweight, binarization-friendly operations. In addition, we propose a Distribution-Aware Binarized Convolution (DAB-Conv) that leverages full-precision activation statistics to alleviate quantization-induced errors caused by diverse and asymmetric feature distributions. Extensive experiments across low-light enhancement, denoising, deblurring, and super-resolution demonstrate that BinRVR achieves a favorable accuracy-efficiency trade-off, reducing computation and parameters by approximately 96\% while incurring only a minor performance degradation compared with full-precision models.

Beyond the strictly binarized setting, the proposed framework extends to multi-bit quantization through a unified distribution-aware formulation, enabling flexible adaptation to different hardware constraints without modifying the core architecture. Moreover, we show that the improved restoration quality provided by BinRVR translates into consistent gains in downstream video applications, including object detection and monocular depth estimation, highlighting its practical value beyond low-level reconstruction. Nevertheless, large-motion deblurring cases may still be challenging, because the recoverable RAW signal can be limited and the lightweight temporal interaction in BIIM avoids heavy motion alignment for efficiency. Future work may explore adaptive bit-width allocation, degradation-adaptive and motion-aware modeling, and real-device deployment optimization to further improve robustness and practical efficiency.

\bibliographystyle{IEEEtran}
\bibliography{reference.bib}

\begin{IEEEbiography}[{\includegraphics[width=1in,height=1.25in,clip,keepaspectratio]{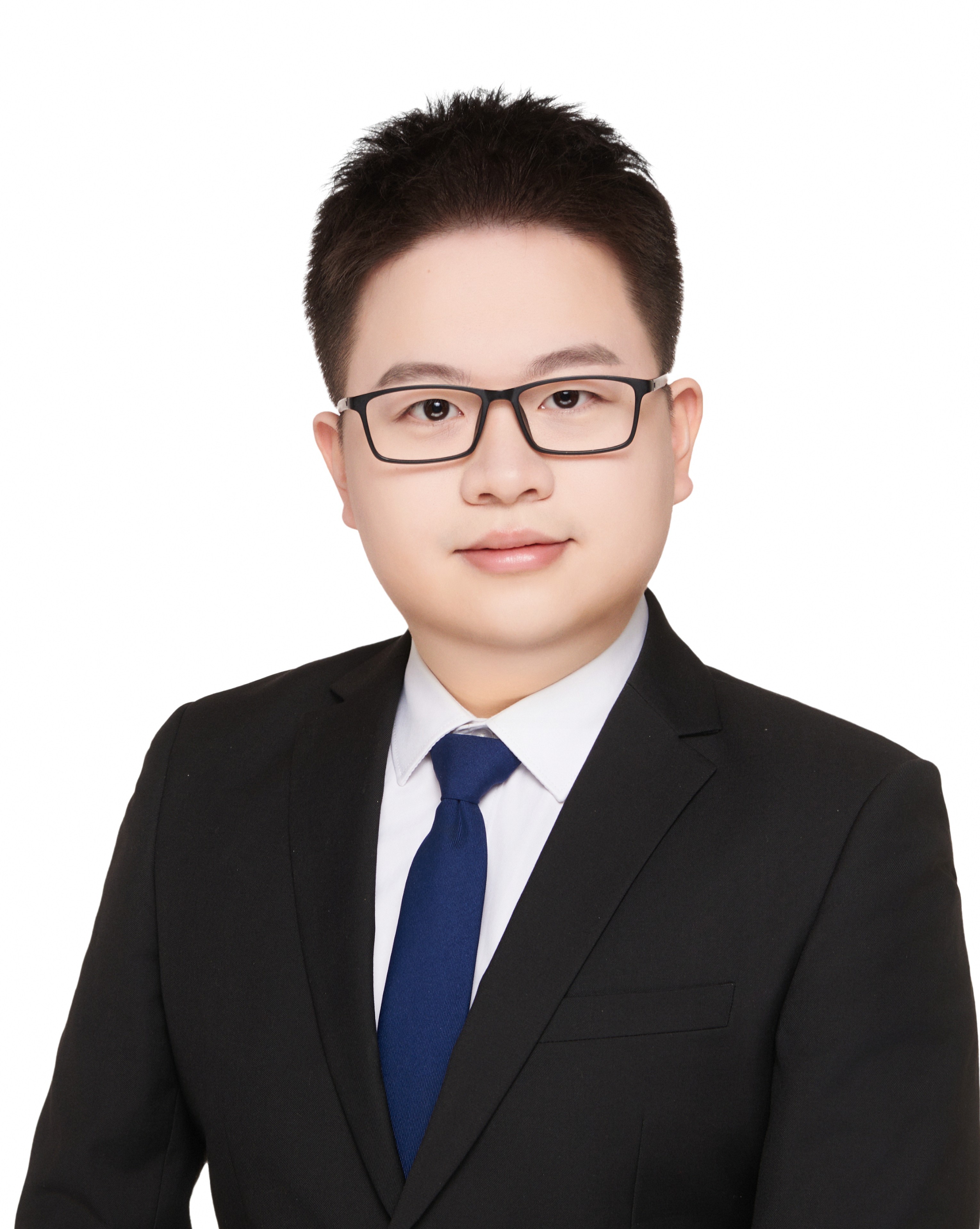}}]{Tianyu Zhu}
received a B.S. degree in computer science and technology from Beijing Institute of Technology, Beijing, China, in 2024. He is currently a M.S. candidate at the School of Computer Science and Technology, Beijing Institute of Technology, Beijing, China. His research interests include artificial intelligence and image processing.
\end{IEEEbiography}
\vspace{-7mm}

\begin{IEEEbiography}[{\includegraphics[width=1in,height=1.25in,clip,keepaspectratio]{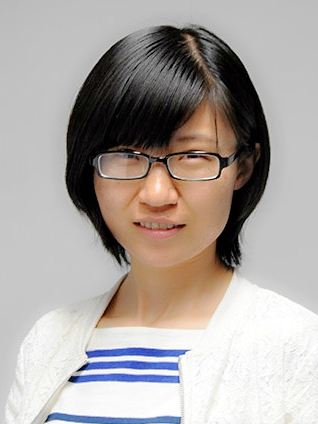}}]{Ying Fu}
received a B.S. degree in electronic engineering from Xidian University, Xi'an, China, in 2009, an M.S. degree in automation from Tsinghua University, Beijing, China, in 2012, and a Ph.D. degree in information science and technology from the University of Tokyo, Japan, in 2015. She is currently a Professor at the School of Computer Science and Technology, Beijing Institute of Technology. Her research interests include computer vision, image and video processing, and computational photography. 
\end{IEEEbiography}
\vspace{-7mm}

\begin{IEEEbiography}[{\includegraphics[width=1in,height=1.25in,clip,keepaspectratio]{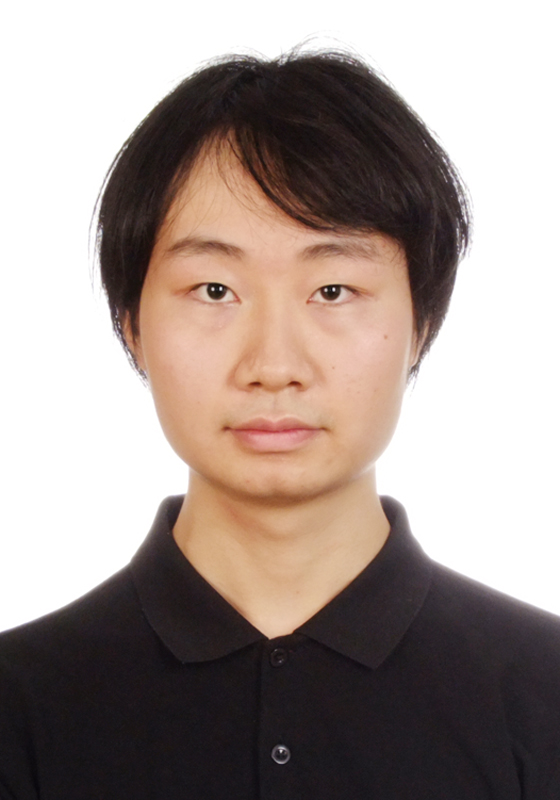}}]{Hesong Li}
received a B.S. degree in mathematics and applied mathematics from Dalian Maritime University, Dalian, China, in 2021. He is currently a Ph.D. candidate at the School of Computer Science and Technology, Beijing Institute of Technology, Beijing, China. His research interests include artificial intelligence and image processing.
\end{IEEEbiography}
\vspace{-7mm}

\begin{IEEEbiography}[{\includegraphics[width=1in,height=1.25in,clip,keepaspectratio]{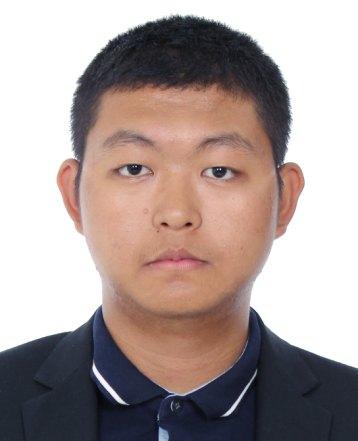}}]{Gengchen Zhang}
Gengchen Zhang received a B.S. degree in computer science from Beijing Institute of Technology, Beijing, China, in 2022, an M.S. degree in computer science from Beijing Institute of Technology, Beijing, China, in 2025.
He is currently working at Chinese Aeronautical Establishment.
His research interests include computational photography, image processing, and deep-learning.
\end{IEEEbiography}
\vspace{-7mm}

\begin{IEEEbiography}[{\includegraphics[width=1in,height=1.25in,clip,keepaspectratio]{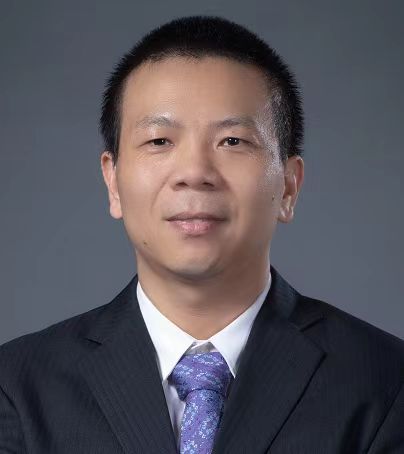}}]{Xin Yuan}
 (SM'16) received the BEng and MEng degrees from Xidian University, in 2007 and 2009, respectively, and the PhD from the Hong Kong Polytechnic University, in 2012. 
He is currently an Associate Professor at Westlake University.  He was a video analysis and coding lead researcher at Bell Labs, Murray Hill, NJ, USA from 2015 to 2021. Prior to this, he was a Post-Doctoral Associate in the Department of Electrical and Computer Engineering, Duke University from 2012 to 2015.
His research interests are 
computational imaging and machine learning. He has been the Associate Editor of {\em Pattern Recognition} (2019-), {\em International Journal of Pattern Recognition and Artificial Intelligence} (2020-) and {\em Chinese Optics Letters} (2021-). He led the special issue of "Deep Learning for High Dimensional Sensing" in the {\em IEEE Journal of Selected Topics in Signal Processing} in 2022.
\end{IEEEbiography}
\vspace{-7mm}

\begin{IEEEbiography}[{\includegraphics[width=1in,height=1.25in,clip,keepaspectratio]{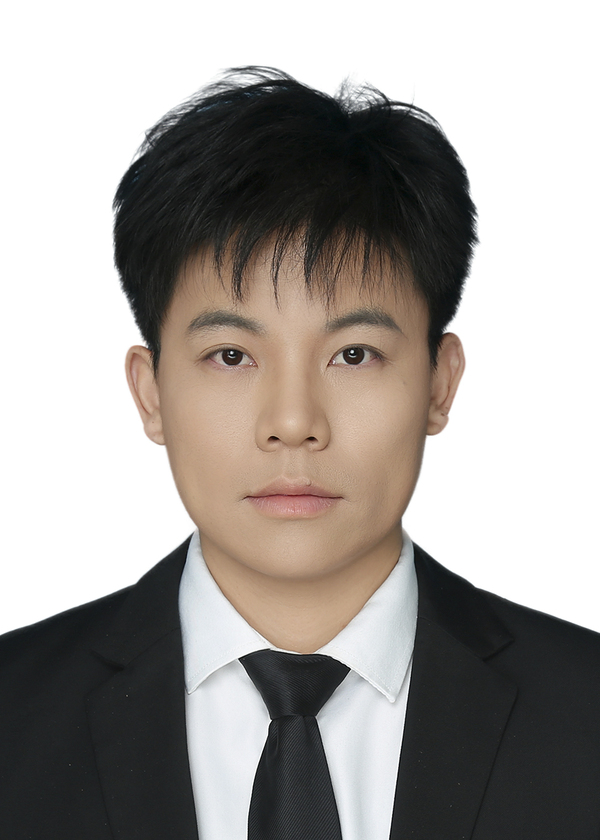}}]{Yulun Zhang}
 received a B.E. degree from the School of Electronic Engineering, Xidian University, China, in 2013, an M.E. degree from the Department of Automation, Tsinghua University, China, in 2017, and a Ph.D. degree from the Department of ECE, Northeastern University, USA, in 2021. He is currently an associate professor at Shanghai Jiao Tong University, Shanghai, China. He was a postdoctoral researcher at Computer Vision Lab, ETH Zürich, Switzerland. He also worked as a research fellow at Harvard University, USA. His research interests include image/video restoration and synthesis, biomedical image analysis, model compression, multimodal computing, large language model, and computational imaging. He is/was an Area Chair for CVPR, ICCV, ECCV, NeurIPS, ICML, ICLR, IJCAI, ACM MM, and a Senior Program Committee (SPC) member for IJCAI and AAAI.
\end{IEEEbiography}

\end{document}